\documentclass[final,5p,times]{elsarticle}
\usepackage{graphics} 
\usepackage{epsfig} 
\usepackage{times} 
\usepackage{amsmath} 
\usepackage{amssymb}  
\usepackage{bm}        
\usepackage{nicefrac}
\usepackage{tikz}
\usepackage{pgfplots}
\pgfplotsset{compat=1.18} 

\usepackage{comment}    
\usepackage{mathrsfs}  
\usepackage[left]{lineno}
\usepackage[mathscr]{euscript}  
\usepackage{graphicx}  
\usepackage{bbold}     
\usepackage{tikz}
\usepackage{amsthm}
\usepackage{comment}
\usepackage{url}
\theoremstyle{remark}
\newtheorem{remark}{Remark}
\usepackage[framemethod=tikz]{mdframed}
\newtheoremstyle{mytheoremstyle} 
  {3pt}   
  {3pt}   
  {\itshape} 
  {}      
  {\bfseries} 
  {}      
  { }     
  {\bfseries #1\ (#3)} 

\theoremstyle{mytheoremstyle}
\newtheorem*{theorem*}{Theorem}

\newmdenv[
  linecolor=black,
  linewidth=1pt,
  backgroundcolor=gray!10,
  innertopmargin=5pt,
  innerbottommargin=5pt,
  skipabove=10pt,
  skipbelow=10pt
]{theorembox}
\usepackage{enumerate}
\DeclareSymbolFont{letters}{OML}{cmm}{m}{it}
\SetSymbolFont{letters}{normal}{OML}{cmm}{m}{it}

\newcommand{\minimize}{\displaystyle\mathop{\mathrm{minimize}}}
\journal{Control Engineering Practice}

\begin{document}
\begin{titlepage}
\centering

This manuscript is currently under review for publication in\\
\textit{Control Engineering Practice}\\[0.5cm]

Iskandar Khemakhem, Manuel Zobel, Johannes Schüle, Oliver Sawodny,\\
Naoki Uchiyama, and Abdallah Farrage\\[0.5cm]

First submission: April 13th, 2026

https://www.sciencedirect.com/journal/control-engineering-practice

\vfill

\end{titlepage}
\begin{frontmatter}
\title{Behavioral Data-Driven Optimal Trajectory Generation for Rotary Cranes}

\author[label1]{Iskandar Khemakhem\corref{cor1}}
\cortext[cor1]{Corresponding author. Email: khemakhem@iams.uni-stuttgart.de}
\author[label1,label2]{Manuel Zobel}
\author[label2]{Johannes Sch{\"u}le}
\author[label2]{Oliver Sawodny}
\author[label3]{Naoki Uchiyama}
\author[label4]{Abdallah Farrage}

\affiliation[label1]{organization={Institute for Adaptive Mechanical Systems, University of Stuttgart},
            city={Stuttgart},
            postcode={70569},
            country={Germany}}

 \affiliation[label2]{organization={Institute for System Dynamics, University of Stuttgart},
            city={Stuttgart},
            postcode={70563},
            country={Germany}}

\affiliation[label3]{organization={Department of Mechanical Engineering, Toyohashi University of Technology},
            city={Toyohashi},
            postcode={441-8580},
            country={Japan}}

\affiliation[label4]{organization={Department of Mechatronics Engineering, Faculty of Engineering, Assiut University,},
            city={Assiut},
            postcode={71515},
            country={Egypt}}
\begin{abstract}
With the growth of the construction industry and the global shortage of skilled labor, the automation of crane control has become increasingly important for safe and efficient operations. 
A central challenge in automatic crane control is the reduction of load oscillations during motion, which is primarily addressed through appropriate slewing trajectories.\\
In this context, classical model-based control methods rely on accurate dynamical models and expert tuning, and often struggle to meet safety and precision requirements, while many learning-based approaches require large data sets and significant computational resources. This paper proposes a behavioral data-driven framework for generating open-loop slewing trajectories for rotary cranes that suppress load sway while reducing operation time and energy consumption.\\
The approach builds on Willems’ fundamental lemma and its generalizations, to bypass explicit system modeling and operate directly on measured input–output data. 
A practical workflow is presented in this paper to reduce the need for expert knowledge.
Despite the underactuated nature of the crane dynamics, the method identifies a nonparametric representation of the system behavior and generates smooth, optimal trajectories using limited data and convex optimization. 
The proposed trajectory generation method is validated on a laboratory crane setup and compared against an established model-based approach, achieving up to~$35\,\%$ reduction in load sway,~$43\,\%$ reduction in tracking error, and~$50\,\%$ reduction in travel time.

\end{abstract}

%
%
%
%

%
%
%
%
%


\end{frontmatter}

\section{Introduction}\label{introduction}
To this day, the construction industry remains one of the least automated sectors~\cite{regina_lauer_state_2023}. 
This slow adoption is mainly due to safety concerns, since most construction sites are located in urban environments and populated with human workers. 
At the heart of this industry are cranes.
They are essential for moving heavy loads, making them central to both safety and efficiency. 
Among the various types of cranes, rotary cranes are the most common in construction sites~\cite{shapira2007cranes}.
With a small footprint, considerable height, and three main degrees of freedom, slewing, luffing, and hoisting, they can efficiently cover a large workspace around the construction area.

However, when the crane boom rotates horizontally, the suspended load tends to sway in both radial and tangential directions.
Human operators typically suppress radial sway by hoisting the load and reduce tangential sway by performing a smooth and slow slewing motion~\cite{takahashi_sensor-less_2022}.
This approach is inefficient, since regulating the slewing motion requires highly skilled operators, and lifting the load lead to unnecessary motion against gravity, resulting in energy and time losses.
Concurrently, the shortage of skilled labor and the rising demand for new infrastructure have created an acute need for novel and safe automation techniques that can be deployed on real crane systems.

To enable safe automation of crane operations, control strategies must mitigate both radial and tangential load sway by adapting the slewing motion during operation. However, full automation based on feedback control remains difficult to achieve and, in practical settings, undesirable \cite{bonnabel2020industrial}. 
Reliable feedback would require precise measurement of the load position, which is difficult to obtain in construction environments where loads are suspended by long, flexible cables. 
In addition, nonlinear and strongly coupled oscillatory dynamics, together with disturbances such as wind, can cause feedback control actions to inject energy into the system, potentially amplifying payload motion and rendering it unstable. 

As a result, crane control predominantly relies on open-loop trajectory planning based on physical models and conservative constraints.
While such approaches offer predictable behavior, their performance is limited by the modeling inaccuracies.
Developing accurate models requires expert knowledge and still fails to capture the system complexity.
These inaccuracies can result in missed targets, inefficient energy use, and a higher risk of accidents, particularly during open-loop tracking.
The increasing availability of data across various domains has therefore driven a growing interest in data-driven control methods that aim to mitigate the effects of model mismatch~\cite{hou_model-based_2013}.

This work proposes a data-driven approach for automating crane slewing motions based on behavioral systems theory.
Using measured operational data, optimal open-loop trajectories are generated by solving a convex optimization problem.
The crane is represented through its observed input–output behavior, avoiding the need for detailed parametric modeling.
This enables the generation of slewing trajectories that reduce radial and tangential load sway while respecting safety constraints.
By operating directly on data, the approach avoids nonlinear and nonconvex optimization problems as well as complex modeling assumptions.

\subsection{Related work}\label{related_work}
Most of the literature on suppressing load-sway in cranes use a model-based approach.
In feedback control, Wolff et al. proposed a nonlinear MPC controller~\cite{wolff_nonlinear_2022}, while Arnold et al. used optimal control to generate a trajectory tracked via a nonlinear controller~\cite{arnold_trajectory_2007}.
Ouyang et al. presented a Lyapunov-based controller~\cite{ouyang_anti-sway_2010}, and proposed generating an S-curve trajectory to control load-sway using horizontal-boom motion~\cite{ouyang_s_curve_2011}.
Using open-loop control, Farrage et al. applied the~A* algorithm to generate obstacle-avoiding, sway-suppressing trajectories~\cite{farrage_trajectory_2023}.
Takahashi et al. proposed a sensor-less time-optimal control for load-sway and boom-twist suppression~\cite{takahashi_sensor-less_2022}.

Prior work has also considered using data-driven method for crane control. 
Bao et al. proposed a data-driven model predictive control approach for crane systems, in which local linear dynamics are learned from closed-loop data using Bayesian optimization \cite{bao2020data}.
Kim et al. explored data-driven modeling for overhead cranes using neural-network–based input–output representations and combined them with adaptive predictive control to achieve swing suppression and accurate positioning \cite{kim2022data}. 
More recently, Chen et al. eliminated the reliance on a physical model by using input–output data within a sliding-mode control framework for overhead cranes, replacing the switching term to mitigate chattering and to incorporate an output-based disturbance observer to ensure robustness in closed-loop control~\cite{chen2025data}.

Within the input-output formulation, Willems’ fundamental lemma provides a foundation for data-driven system representations by characterizing the behavior of linear time-invariant (LTI) systems directly from measured trajectories~\cite{willems_note_2005}.
The lemma provides a powerful identification tool for linear systems, proving that even a seconds-long data trajectory can capture the system dynamics.
Additionally, it forms the basis for multiple novel methods in optimal control~\cite{berberich2020trajectory}.
To name a few, Berberich et al. extended linear-MPC to the behavioral setting and provided stability and robustness guarantees~\cite{berberich_data-driven_2021}. Coulson et al. presented a trajectory tracking optimal control formulation, proved its equivalence to MPC for linear systems, and proposed adding a regularization for nonlinear systems~\cite{coulson_data-enabled_2019}.

These control strategies have shown remarkable results in controlling nonlinear systems, despite the theory being exclusive for linear time-invariant systems~\cite{carron_data-driven_2019, elokda_data-enabled_2021, markovsky_data-driven_2023}.
In this context, Markovsky et al. introduced a generic optimization problem that used the nonparametric representation resulting from the fundamental lemma to handle classical signal processing and control tasks like simulation, smoothing, and prediction~\cite{markovsky_data-driven_2022, markovsky2021behavioral}.
Using their method, performing such tasks only requires solving a convex optimization problem followed by simple algebraic computations.

\subsection{Contribution}\label{contribution}
This work uses behavioral systems theory to generate optimal trajectories for tower cranes.
To the best of our knowledge, this is the first time behavioral systems theory is employed in crane control.
By operating directly on measured data, our approach enables accurate tracking with reduced radial and tangential load sway, while requiring lower computational effort than model-based alternatives.
These properties allow faster trajectory generation and more reliable open-loop execution under realistic measurement noise, addressing key practical limitations of existing approaches.

We validate our approach on an experimental crane and present a practical workflow that allows the proposed framework to be readily applied to other crane systems.
Additionally, we compare our results with a classical model-based optimal trajectory generation formulation.
Our method achieved more accurate predictions and generated more effective trajectories, resulting in rapid crane motion with lower load oscillation.

Methodologically, our work is most inspired by \cite{markovsky_data-driven_2022} and~\cite{markovsky2021behavioral}, and goes one step further by extending the proposed generic optimization problem to generate smooth optimal trajectories for nonlinear dynamics directly from noisy data to interpolate between given start and desired target conditions while adhering to output constraints and time constraints. 
By eliminating the need for explicit parameter identification and nonlinear ODE solvers, the proposed method significantly simplifies the open-loop control pipeline. 
It nevertheless requires careful data collection, tuning of hyperparameters, and solving a convex optimization problem. 
Owing to its conceptual simplicity, the method can in principle be applied to other systems with limited additional expert knowledge. But a more extensive experimental evaluation is left for future work.\\

The remainder of this article is organized as follows. Section~\ref{chap_setup} introduces the experimental setup and the system dynamics used throughout the study. Section~\ref{chap_trajGen} outlines the theoretical foundations and presents the proposed trajectory generation method. Section~\ref{chap_practical_workflow} formulates the data-driven trajectory generation problem and describes a practical approach for generating the required data under realistic conditions. This is followed by an experimental validation, including a comparison with a benchmark model-based optimization technique, in Section~\ref{chap_experimental}. Finally, the article concludes in Section~\ref{chap_conclusion}.

\section{Experimental Setup}\label{chap_setup}
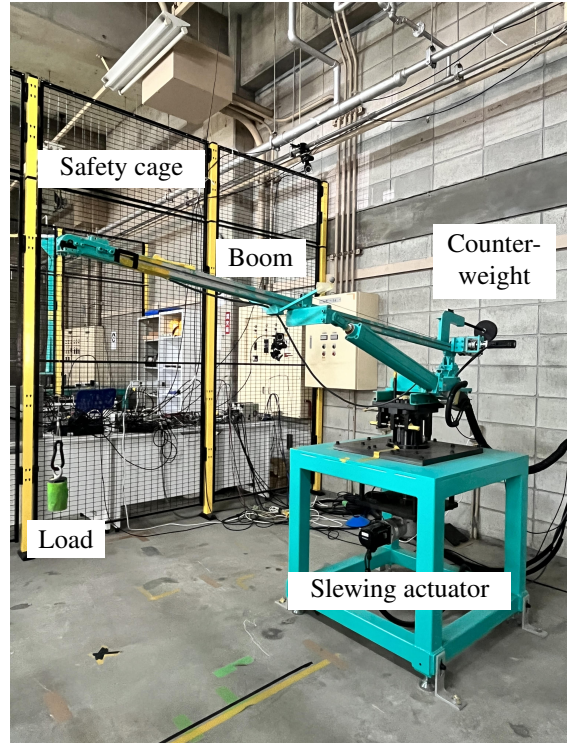
\begin{figure}
	\centering
	\begin{tikzpicture}
	\node [anchor=south west, inner sep=0] (image) at (0,0) {\includegraphics[width=0.4\textwidth]{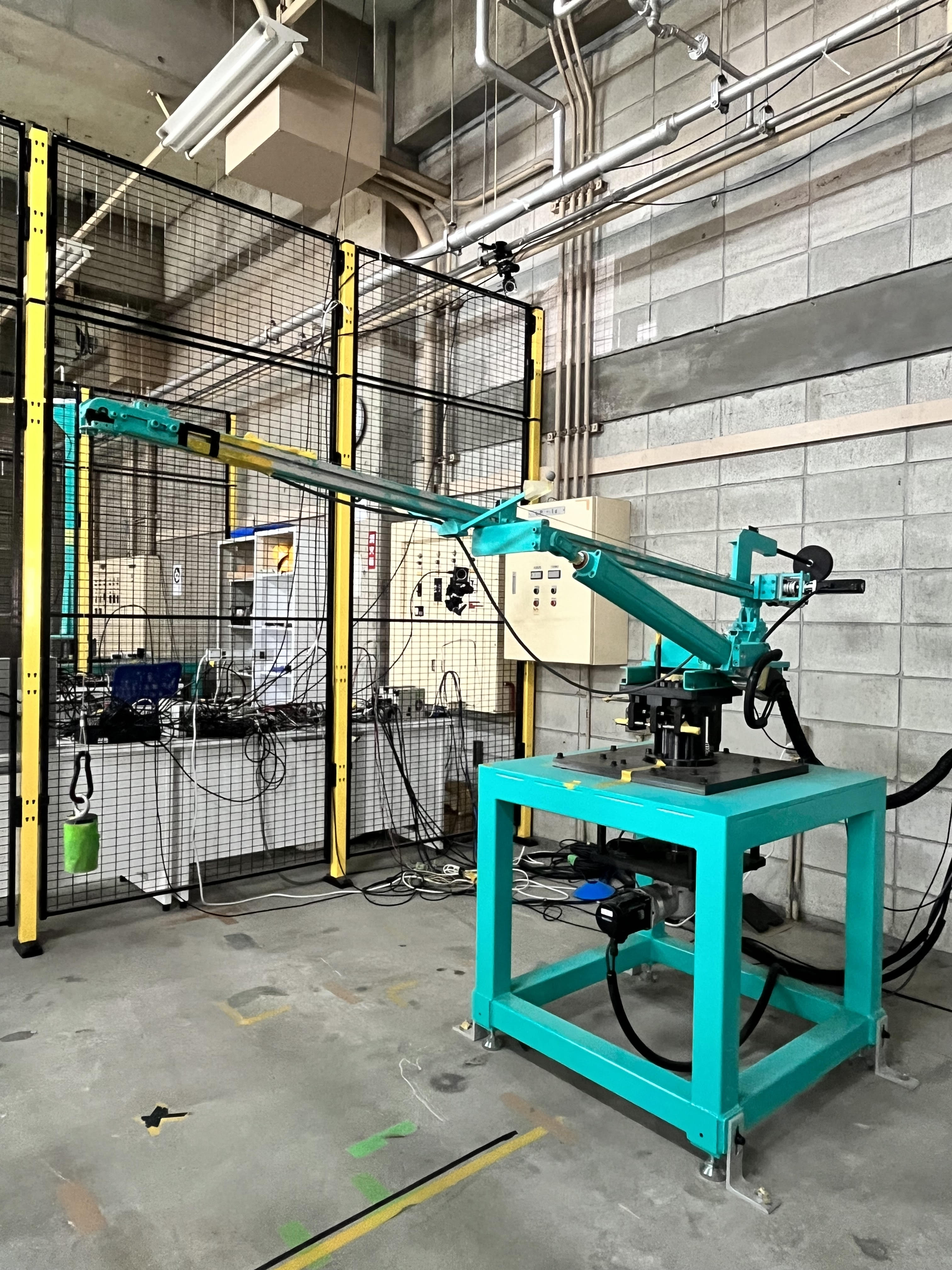}};

	\begin{scope}[x={(image.south east)}, y={(image.north west)}]

	\draw[fill=white, draw=none] (0.5, 0.18) rectangle (0.9, 0.23);
	\node[align=center] at (0.7, 0.205) {Slewing actuator};

	 \draw[fill=white, draw=none] (0.03, 0.25) rectangle (0.17, 0.3);
	\node[align=center] at (0.1, 0.275) {Load};

	\draw[fill=white, draw=none] (0.76, 0.6) rectangle (0.98, 0.7);
	\node[align=center] at (0.87, 0.65) {Counter-\\weight};

	 \draw[fill=white, draw=none] (0.37, 0.68) rectangle (0.53, 0.63);
	\node[align=center] at (0.45, 0.655) {Boom};

	\draw[fill=white, draw=none] (0.05, 0.8) rectangle (0.35, 0.75);
	\node[align=center] at (0.2, 0.775) {Safety cage};

	\end{scope}
\end{tikzpicture}
	\caption{The laboratory rotary crane used in this work.}
	\label{fig_lab_crane_image}
\end{figure}

Experiments in this work are conducted on a laboratory rotary crane, depicted in Fig.~\ref{fig_lab_crane_image}. 
The crane represents a 1:9 down-scaled version of a commercial tower crane.
It is mounted on a stationary platform and rotates around a vertical axis using an AC motor located underneath the platform. 
The crane boom of length~$l_b = 2\,\text{m}$ rotates only about the vertical axis, with its luffing angle fixed.
The load weighs $1\,\text{kg}$ and is suspended by a cable of length $l = 1\,\text{m}$ wrapped around a drum. 
A counterweight is installed on the back of the boom for balance.
The experimental crane is located inside a building in a safety cage. 
Hence, handling wind disturbances and unknown objects are excluded in this study.

\begin{figure}
\centering
\includegraphics[width=0.5\textwidth]{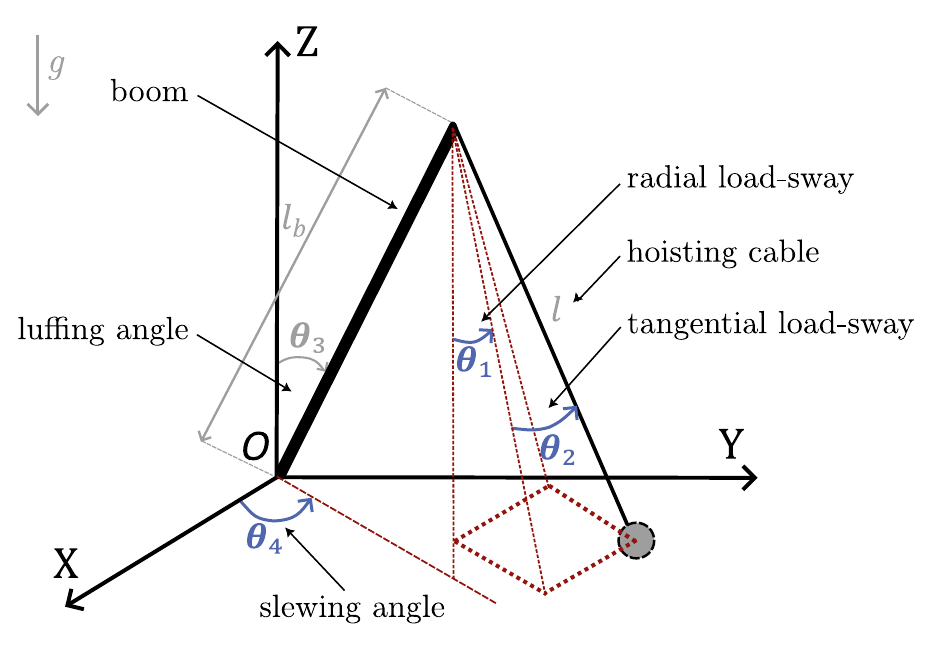}
\caption{Schematic model of a rotary crane with the crane coordinates and parameters.
The variables~$\theta_1$,~$\theta_2$, and~$\theta_4$ are depicted in blue.
The luffing angle~$\theta_3$ and the cable length~$l$ are considered constant and depicted in grey similar to the remaining parameters.}
\label{fig_crane_coordinates}
\end{figure}

Fig.~\ref{fig_crane_coordinates} presents a schematic model of a rotary crane where~$\theta_1$ and~$\theta_2$ represent the radial and tangential load sway angles,~$\theta_3$ and~$\theta_4$ are the luffing and slewing boom angles,~$l_b$ the boom length,~$l$ the cable length, and~$g$ the gravitational constant.
Table~\ref{tab:exp_params} summarizes the parameters of the experimental setup.

The laboratory crane system is velocity-controlled. 
A digital signal processor sends real-time velocity commands to a servo-controller.
The servo-controller transforms the slewing angular velocities~$\dot{\theta}_4$ into motor voltages using a proportional term.
A simple visual system tracks the position of the load and the boom positions with a sampling rate of~$20\,\text{Hz}$.
Inverse kinematics are used to extract the load angles~$\theta_1$ and~$\theta_2$, as well as the boom slewing angle $\theta_4$.

The goal of this work is to move the load from a certain initial position to a given target position. 
Given that the luffing angle and hoisting cable are fixed, the load initial position will correspond to an initial slewing angle $\theta^\mathrm{start}_4$ and target angle $\theta^\mathrm{target}_4$ in the absence of load-sway.

\begin{table}[t]
    \centering

    \begin{tabular}{l c c c}
        \hline
        \textbf{Parameter} & \textbf{Symbol} & \textbf{Value} \\
        \hline
        Boom length     & $l_b$ & $2\,\text{m}$ \\
        hoisting cable length     & $l$  & $1\,\text{m}$ \\
        load weight      & - & $1\,\text{kg}$  \\
        luffing angle & $\theta_3$  & $\frac{\pi}{4}\,\text{rad}$ \\
        highest slewing angular velocity & $|\dot{\theta}^\text{max}_4|$  & $0.6\,\text{rad/s}$ \\
        \hline
    \end{tabular}
    \caption{Summary of experimental step parameters    \label{tab:exp_params}}
\end{table}

%
%
%

%
%

%
%
%
%
%
%
%
%
%
%
%

%
%

%
%
%
%
%

\section{Trajectory Generation using Behavioral Systems Theory}\label{chap_trajGen}
This section presents the main methodological contribution of this paper. 
A brief introduction to the fundamentals of behavioral systems theory is presented in Sections~\ref{chap_fundamentals} and~\ref{sec_missing_data_recovery}, which summarize the key results from~\cite{markovsky_data-driven_2023}, ~\cite{markovsky_identifiability_2023},~\cite{Coulson_distributionally} and~\cite{dorfler_bridging_2023}. 
The main methodological contribution of this work builds upon these results, which will be further developed in Section~\ref{chap_optimal_traj}.

\subsection{Fundamentals of Behavioral Systems Theory}\label{chap_fundamentals}

In behavioral systems theory, a system is fully characterized by its behavior, that is, the set of all trajectories it can exhibit~\cite{willems1986time}. 
In this context, a trajectory~$\bm{w} \in (\mathbb{R}^{q})^N$, with~$N \in \mathbb{N}$ refers to the combination of an~$N$-long sequence of input samples and their resulting output samples, all assembled in a vector in a user-defined order. 
The vector~$\bm{w}$ represent the time-discrete trajectory of a dynamical system.
We denote the number of elements in a sample with~$q$ and the number of inputs with~$m$. 
The sampling frequency is assumed to be always constant.

For the system considered in this work, an input-output sample for the experimental crane is given as
\begin{equation}\label{eq_trajectory_sample}
     \bm{w}(i) = \begin{bmatrix}
     \dot{\theta}_4(i) & \theta_1(i) & \theta_2(i) & \theta_4(i)
     \end{bmatrix}^\text{T},
\end{equation}
with~$\dot{\theta}_4$ representing the system's input. 
For the crane's system~$q = 4$ and~$m = 1$. 
Throughout this work, the sampling frequency is kept at~$20\,\text{Hz}$.

We define the truncation operator~$(\phantom{w})|^\mathcal{I}$, that retains the elements of a vector or the rows of a matrix specified by~$\mathcal{I}$, i.e, for~$\mathcal{I} = \{j_1, j_2, \dots, j_d\}$ and a vector~$\bm{w}$,
\begin{equation}
    \bm{w}|^\mathcal{I} = \begin{bmatrix}
        \bm{w}[j_1]& \bm{w}[j_2] & \cdots& \bm{w}[j_d]
    \end{bmatrix}^\text{T}.
\end{equation}
Here, we distinguish between a \emph{trajectory sample}~$\bm{w}(i)$, which collects all input and output signals at time~$i$, and a \emph{trajectory element}~$\bm{w}[j]$, which denotes the~$j$-th scalar component of the vector~$\bm{w}$.

The behavior of a dynamical system is a subspace~$\mathcal{B}$ in the space of trajectories, that is, in the space of infinitely long sampled trajectories,~$\mathcal{B}$ contains all the trajectories that can be observed from a dynamical system . 
In particular, finite-length trajectories of an LTI system generate a linear and shift-invariant subspace, that is 
\begin{equation}
    \forall\;\bm{w}_1, \bm{w}_2 \in \mathcal{B} \; \text{and} \; \forall\, \alpha, \beta \in \mathbb{R},\; \alpha \bm{w}_1 + \beta \bm{w}_2 \in \mathcal{B} 
\end{equation}
and,
\begin{equation}
    \forall\;\bm{w}_1, \bm{w}_2 \in \mathcal{B}, \quad \mathcal{B} \subseteq \Gamma \mathcal{B}
\end{equation}
with \mbox{$\Gamma\mathcal{B} \coloneqq \{\Gamma\bm{w}\;|\;\bm{w}\in\mathcal{B}\}$} and~$\Gamma\bm{w}(i) \coloneqq \bm{w}(i+1)$~\cite{willems1986time}.

When considering trajectories of finite length~$N$, the restricted behavior of the dynamical system~$\mathcal{B}$ in the set~$\{1,\dots, N\}$ is defined as~$\mathcal{B}|^N \coloneqq \{\bm{w} \in (\mathbb{R}^{q})^N \;|\; \exists \bm{v} \in \mathcal{B}\,: \bm{v}|^{\mathcal{I}_i} = \bm{w}|^{\mathcal{I}_i} \forall i \in \{1,\dots, N\} \}$, with the set~$\mathcal{I}_i$ containing all samples up to the~$i$-th sample~\cite{padoan2022behavioral}.

The order~$n$ of a system denotes the dimension of its minimal state-space realization and corresponds to the minimum number of internal states required to describe its dynamics. The lag~$\ell$ characterizes the dependencies among consecutive samples of a discrete-time trajectory and can be interpreted as the number of past samples required to uniquely reconstruct the state in minimal state-space realization. More formal definitions of the lag can be found in \cite{coulson_data-enabled_2019} and \cite{markovsky2006exact}.
In behavioral systems theory, we assume that only the number of inputs~$m$ and outputs~$p$ are known, where~$p = q - m$. All other system parameters, including the order~$n$ and lag~$\ell$, are considered unknown.

Let~$\bm{w}^d$ be a data trajectory of length~$N$ generated from an LTI system of order~$n$ and lag~$\ell$.
For any~$L < N$, the vector~$\bm{w}^d$ is organized in a Hankel structure of depth~$L$, 
\begin{equation}\label{eq_Hankel_matrix}
 \bm{\mathcal{H}}_L(\bm{w}^d) := 
\begin{bmatrix}
 \bm{w}^d(1) & \bm{w}^d(2) & \cdots & \bm{w}^d(N-L+1) \\
 \bm{w}^d(2) & \bm{w}^d(3) & \cdots & \bm{w}^d(N-L+2) \\
 \vdots & \vdots & \ddots & \vdots \\
 \bm{w}^d(L) & \bm{w}^d(L+1) & \cdots & \bm{w}^d(N)
\end{bmatrix}.
\end{equation}
The Hankel structure in Eq.~\eqref{eq_Hankel_matrix} uses the shift-invariance property of LTI systems to extract the maximum number of valid trajectories of length~$L$ from a trajectory of length~$N$ and organizes them as columns in a matrix. A similar structure can be used when multiple data sequences are available from the system, where all Hankel submatrices are concatenated to create a matrix of trajectories~\cite{van_waarde_extension}. 
For clarity, this paper uses the notation of a single data sequence to represent the entire set of generated data, even when multiple sequences are used.

For a given trajectory of noise-free data~$\bm{w}^d$ of~$N$ samples, generated from an LTI system~$\mathcal{B}$ of order~$n$, lag~$\ell$ and~$m$ inputs, and for any~\mbox{$L>\ell$},~$\bm{\mathcal{H}}_{L}(\bm{w}^d)$ spans the restricted behavior~$\mathcal{B}|^L$ if and only~if
\begin{equation} \label{Eq_PE}
 \text{rank}\, \bm{\mathcal{H}}_{L}(\bm{w}^d) = mL+n,
\end{equation}
i.e., for any~$L$-long trajectory~$\bm{w}$ of~$\mathcal{B}|^L$, there exists a vector~$\textbf{\textsl{g}}~\in~\mathbb{R}^{N-L+1}$, such that
\begin{equation}\label{eq_lemma_w}
 \bm{w} = \bm{\mathcal{H}}_{L}(\bm{w}^d)\textbf{\textsl{g}}.
\end{equation}
This is a generalization of Willems' fundamental lemma~\cite{markovsky_identifiability_2023}.
It leverages the Hankel structure and the superposition principle in linear subspaces to provide a nonparametric representation of the behavior of an LTI system in Eq.~\eqref{eq_lemma_w}.
The condition in Eq.~\eqref{Eq_PE} is called the identifiability condition.

\subsection{Missing Data Recovery Problem}\label{sec_missing_data_recovery}
\subsubsection{Recovering Data from Noise-free Linear Dynamics}\label{sec_missin_data_recovery_linear}
Suppose there exists a finite sequence~$\bm{w}^d$ of~$\mathcal{B}|^N$ that satisfies Eq.~\eqref{Eq_PE}.
Further, assume that some elements of a trajectory~$\bm{w}\in\mathcal{B}|^L$ are missing. 
Let the set~$\mathcal{I}$ contain~$d < qL$ distinct indices~$j_i \in \{1,\hdots,qL\}$ denoting the known elements of the trajectory~$\bm{w}$. 
By solving for \textbf{\textsl{g}} the linear system of equations
\begin{equation}\label{eq_lin_system}
 \bm{w}|^{\mathcal{I}} = \bm{\mathcal{H}}_L(\bm{w}^d)|^{\mathcal{I}} \textbf{\textsl{g}},
\end{equation}
the trajectory~$\bm{w}$ can be fully recovered using Eq.~\eqref{eq_lemma_w}.

\begin{figure}
   \centering
   \includegraphics[width=0.5\textwidth]{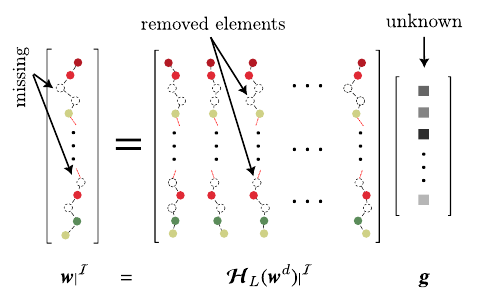} 
   \caption{Schematic of the data-driven trajectory interpolation problem: 
   On the left,~$\bm{w}|^{\mathcal{I}}$ shows a trajectory~$\bm{w}$ with the missing points in white dashed circles. 
   On the right, the trajectory submatrix~$\bm{\mathcal{H}}_L(\bm{w}^d)|^{\mathcal{I}}$, with the rows for missing points removed, and the unknown weight vector~$\textbf{\textsl{g}}$ are shown. 
   The vector~$\textbf{\textsl{g}}$ remains the same as in the full trajectory.
   The goal is to solve this system for~$\textbf{\textsl{g}}$ to recover the full trajectory~$\bm{w}$.}
   \label{fig_interpolation}
\end{figure}

Recovering the missing data in~$\bm{w}$ using the Hankel matrix represents the data-driven interpolation introduced in~\cite{markovsky_data-driven_2022}.
The idea behind this method is that truncating~$\bm{w}$ and the rows of~$\bm{\mathcal{H}}_L(\bm{w}^d)$ to~$\mathcal{I}$ does not alter~$\textbf{\textsl{g}}$.
Fig.~\ref{fig_interpolation} illustrates Eq.~\eqref{eq_lin_system}.

In case the given trajectory~$\bm{w}|^{\mathcal{I}}$ is not consistent with the data in~$\bm{\mathcal{H}}_L(\bm{w}^d)|^{\mathcal{I}}$, a solution for Eq.~\eqref{eq_lin_system} does not exist. 
Replacing Eq.~\eqref{eq_lin_system} by the approximation problem
\begin{equation} \label{eq_opt_approximation}
 \minimize_{\textbf{\textsl{g}}} \| \bm{w}|^{\mathcal{I}} - \bm{\mathcal{H}}_L(\bm{w}^d)|^{\mathcal{I}}\textbf{\textsl{g}}\|_\mathbf{W}^2,
\end{equation}
with the weighted Euclidean norm~$\|\bm{v}\|_\mathbf{W}^2 = \bm{v}^\mathrm{T}\mathbf{W}\bm{v}$,
finds the closest solution~$\hat{\bm{w}}\in\mathcal{B}|^L$ to the trajectory with missing elements~$\bm{w}$.
The weight matrix~$\bm{W}$ enables different weighting of the known elements according to their relative importance, in the approximation.
Eq.~\eqref{eq_opt_approximation} followed by Eq.~\eqref{eq_lemma_w} represent the generic data-driven interpolation and approximation problem for the restricted behavior of an LTI system identified using noise-free data presented in~\cite{markovsky_data-driven_2022}. 

If~$\bm{\mathcal{H}}_L(\bm{w}^d)|^{\mathcal{I}}$ has full column rank, then Problem~\eqref{eq_opt_approximation} admits a unique solution.
However, in some cases infinitely many valid trajectories exist, whose truncations to~$\mathcal{I}$ coincide with~$\bm{w}$. 

\subsubsection{Extension to Nonlinear Dynamics}\label{sec_nonlinear}
Most real systems, including the rotary crane, are nonlinear.
In classical settings, nonlinear dynamics can be locally approximated by linear ones via Taylor series expansions, whereas the approximation is only valid near a specific point.
Using Koopman theory~\cite{proctor2018generalizing} or Carleman linearization~\cite{amini2025carleman}, a nonlinear system may also be lifted to a higher-dimensional space where it is approximately linear on a finite time horizon.

While applying the fundamental lemma to nonlinear systems has shown remarkable success, to this day the nonlinear fundamental lemma is only proven for specific classes of systems, such as feedback-linearizable nonlinear systems~\cite{alsalti2023data}, bilinear systems~\cite{yuan2022data}, as well as Hammerstein and Wiener systems~\cite{berberich2020trajectory}.
The link between applying the fundamental lemma to nonlinear systems and lifting the system using Koopman theory has been established for several years (e.g., in~\cite{berberich2020trajectory}). A recent preprint presents a proof for the existence of an infinite-dimensional bilinear Koopman embedding for nonlinear systems with input and exploits it to derive a nonlinear fundamental lemma~\cite{lazar2025product}. 

This work does not explicitly follow any of the mentioned formalisms. Instead, it draws inspiration from ideas in Koopman lifting and Taylor series expansions to build intuition for the forthcoming framework. 
In particular, we interpret the Hankel matrix as a dictionary of motion primitives, where each subset of columns corresponds to a local linear approximation of the underlying nonlinear dynamics in a region around a specific operating point.
Additionally, by lifting the system dynamics, that is, by increasing the effective order~$n$ of the representation, the finite-horizon behavior of the system can be locally approximated by linear dynamics. 
Through careful design of the input sequence used for data generation, it is therefore possible to target specific regions of the state space in which this approximation remains valid, while residual nonlinear effects can be treated as noise.

The size of the captured region of the dynamics is directly related to the amount and diversity of the collected data. 
Generating additional data increases the number of columns in the Hankel matrix, which in turn augments the number of coefficients in the vector~$\textbf{\textsl{g}}$ and consequently enlarges the dimension of the search space in Problem~\eqref{eq_opt_approximation}. 
In practice, however, the system’s motion typically evolves within a low-dimensional subspace of the captured dynamics.
This observation motivates the introduction of a regularization term to promote the selection of the relevant motion primitives from the dictionary encoded in~$\bm{\mathcal{H}}_L(\bm{w}^d)$. 
To this end, we augment Eq.~\eqref{eq_opt_approximation} with an~$\ell_1$-norm regularization term on the vector~$\textbf{\textsl{g}}$, as proposed in~\cite{markovsky_data-driven_2022}, leading to the problem:
\begin{equation}\label{eq_opt_lasso}
    \minimize_{\textbf{\textsl{g}}} \| \bm{w}|^{\mathcal{I}} - \bm{\mathcal{H}}_L(\bm{w}^d)|^{\mathcal{I}}\textbf{\textsl{g}}\|_\mathbf{W}^2 + \lambda \|\textbf{\textsl{g}}\|_1.
\end{equation}

The~$\ell_1$-norm is a convex regularization term that promotes sparsity in the vector~$\textbf{\textsl{g}}$. 
When combined with Eq.~\eqref{eq_lemma_w}, this sparsity effectively enforces the selection of a subset of columns from the Hankel matrix~$\bm{\mathcal{H}}_L(\bm{w}^d)$. 
As a result, the optimization problem implicitly performs feature selection over the columns of~$\bm{\mathcal{H}}_L(\bm{w}^d)$.

This formulation avoids any explicit projection onto a linear model space~\cite{dorfler_bridging_2023}, and hence is considered as a direct data-driven approximation and interpolation technique for nonlinear systems.
Its performance is generally superior to that of indirect data-driven methods, in which the data are first projected onto the set of LTI behaviors~\cite{markovsky_data-driven_2022}.
Moreover, the sparse regularization adds an~$\ell_\infty$ distributional robustness to the optimization problem~\cite{Coulson_distributionally}.  

\vspace{1em}
\begin{remark}
    \ref{sec_controllability_observability} present a model-based analysis of the crane's controllability. When controlled only through the slewing motion, the crane is only nonlinearly accessible. This class of systems are in fact commonly controlled using motion primitives as other control and planning methods are non-intuitive.
\end{remark}

\subsubsection{Extension to Noisy Data}\label{sec_noisy_data}
Data generated from a real system are inevitably affected by noise.
Adding a random noise to two linearly dependent vectors almost always makes them linearly independent, thus increasing the rank of~$\bm{\mathcal{H}}_L(\bm{w}^d)$. 
Consequently, Eq.~\eqref{Eq_PE} becomes trivially satisfied when~$mL+n$ columns are available, making it impractical to verify the system identifiability using Eq.~\eqref{Eq_PE}.

When the noise magnitude is small relative to the actual data, the essential structure of the Hankel matrix is expected to be preserved, with the noise mainly affecting the smallest singular values. 
Consequently, truncating these singular values is commonly used as a heuristic to attenuate the effect of noise and to obtain a reduced-rank approximation of~$\bm{\mathcal{H}}_L(\bm{w}^d)$. 

Given~$\mathrm{rank}\,\bm{\mathcal{H}}_L(\bm{w}^d) = r$ and any~$\bar{r}<r$, the Eckart--Young theorem states that truncating the~$r-\bar{r}$ smallest singular values yields the closest rank-$\bar{r}$ approximation of~$\bm{\mathcal{H}}_L(\bm{w}^d)$ in the Frobenius and spectral norms~\cite{eckart_approximation_1936}.

\subsection{Behavioral Optimal Trajectory Generation}\label{chap_optimal_traj}
In crane control for heavy-load transport, the primary objective of trajectory generation is to drive the system from given initial conditions to desired target conditions while ensuring low load sway for operational safety. 
More generally, optimal trajectory generation seeks to additionally minimize energy consumption and travel time while satisfying system constraints such as input and output bounds. 

This section presents a trajectory generation framework for rotary cranes based on behavioral systems theory, which constitutes the core contribution of the paper.
By leveraging the missing data recovery formulation introduced in Section~\ref{sec_missing_data_recovery}, the proposed approach generates optimal trajectories directly from measured data while satisfying user-defined requirements.

The objective is to compute an~$L$-sample long trajectory that takes the crane from a resting configuration at a given boom angle~$\theta_4^\text{start}$, with the load in its downward equilibrium, to a target resting configuration at~$\theta_4^\text{target}$, with the load again at equilibrium.
Although the trajectory length is fixed to $L$ samples, the crane is required to reach the target configuration as early as possible and remain there thereafter. Simultaneously, the trajectory is optimized to minimize energy consumption.

In the behavioral setting, initial and target resting conditions are manifested by a sequence of identical samples of length~\mbox{$N_\text{given} > \ell$} at the beginning and end of the trajectory~$\bm{w}$. 
Let the set of known elements be defined as,
\begin{equation}
    \mathcal{I} = \{1, \hdots, qN_\text{given},\, qL - qN_\text{given} + 1, \hdots, qL\}.
\end{equation}
Generating a trajectory that interpolates between the start and target is equivalent to recovering~$\bm{w}$ from the partially known trajectory~$\bm{w}|^\mathcal{I}$ using the dynamics captured in~$\bm{\mathcal{H}}_L(\bm{w}^d)$.  
This represents a special case of the missing data recovery problem presented in the previous section.

We seek a trade-off between three competing objectives: minimal load sway, minimal transition time, and minimal energy. 
Load sway is minimized by maintaining the load the closest possible to its downward equilibrium, which is hard to achieve if the boom trajectory is not smooth. 
The time objective is addressed by minimizing the distance to the target boom position, promoting a swift arrival. 
Finally, energy efficiency is achieved by minimizing the squared control effort.

These objectives can be combined into a unified cost function as follows.
Let~$\bm{w}_{\text{ref}}$ be an~$L$-sample-long reference trajectory defined as
\begin{equation}
 \bm{w}_\text{ref} = \left[
 \begin{array}{cccc}
 \underbrace{\bm{w}_\text{start}^\mathrm{T} \ldots \bm{w}_\text{start}^\mathrm{T}}_{N_\text{given}} &
 \underbrace{\bm{w}_\text{target}^\mathrm{T} \ldots \bm{w}_\text{target}^\mathrm{T}}_{L - N_\text{given}}
 \end{array}
 \right]^\mathrm{T}.
\end{equation}
The reference trajectory~$\bm{w}_{\text{ref}}$ jumps after~$N_\text{given}$ samples from the initial to the target conditions, while the input remains zero.
In the crane's case,~$\bm{w}_{\text{ref}}$ features a jump in~$\theta_4$ from~$\theta_4^\text{start}$ to~$\theta_4^\text{target}$, while all other components remain zero.
Minimizing the squared Euclidean distance
\begin{equation}\label{eq_opt_reference}
    \| \bm{w}_\text{ref} - \bm{\mathcal{H}}_L(\bm{w}^d)\,\textbf{\textsl{g}} \|_\mathbf{R}^2,
\end{equation}
yields a solution that approaches the target quickly while keeping inputs and load angles near zero, effectively reducing both sway and energy consumption. 
The weight matrix $\bm{R}$ allows the user to tune the trade-off between the three objectives.

\vspace{1em}
\begin{remark}
In a general setting, i.e., without restricting the discussion to the crane system, the input at the two operating points does not need to be zero. The key requirement is that the corresponding input–output pair forms an equilibrium of the underlying nonlinear system. In our setup, this condition is satisfied only in the special case of zero input.
\end{remark}
\vspace{1em}

To ensure smooth boom motion and avoid abrupt velocity changes that induce load oscillations, we penalize the total variation of~$\theta_4$ and~$\dot{\theta}_4$.
Let~$\bm{\mathcal{D}}$ be a linear operator that computes finite differences of the corresponding selected elements in~$\bm{\mathcal{H}}_L(\bm{w}^d)$.
To favor smooth boom trajectories, we include a regularization term based on the squared norm of the difference of the trajectories of $\theta_4$ and $\dot{\theta}_4$, implemented through the operator 
$\bm{\mathcal{D}}$, which acts on the corresponding rows of the Hankel matrix and applies zero padding at the boundaries,
\begin{equation}\label{eq_opt_total_variation}
   \left\| \bm{\mathcal{D}}\,\bm{\mathcal{H}}_L(\bm{w}^d)\,\textbf{\textsl{g}} \right\|^2.
\end{equation}

Given that the known elements in a trajectory generation task are limited to the initial and final sequences, infinitely many solutions to Eq.~\eqref{eq_lin_system} exist, and consequently, infinitely many trajectories interpolate between the initial and target conditions. 
Solving the convex optimization problem in Eq.~\eqref{eq_opt_lasso} yields a particular solution~$\textbf{\textsl{g}}_p$. 
By incorporating the trajectory optimization objective of Eq.~\eqref{eq_opt_reference} and the regularization term in Eq.~\eqref{eq_opt_total_variation} as weighted terms in the cost function, we force the particular solution~$\textbf{\textsl{g}}_p$ to coincide with the trajectory that fits our requirements the best. 

Since Problem~\eqref{eq_opt_lasso} is an approximation problem, the resulting trajectory might not match the actual crane start and target positions exactly. To address this, we add the following equality constraint to the optimization problem:
\begin{equation}
\bm{\mathcal{H}}_L(\bm{w}^d)|^\mathcal{J}\textbf{\textsl{g}} =
\begin{bmatrix}
\bm{w}_\text{start}^\mathrm{T} \
\bm{w}_\text{target}^\mathrm{T}
\end{bmatrix}^\mathrm{T},
\end{equation}
with $\mathcal{J} = \{1, \hdots, q,\, (q-1)L + 1, \hdots, qL\}$.
Additionally, we impose strict bounds on the input and load sway for safety,
\begin{equation}
\bm{M}\bm{\mathcal{H}}_L(\bm{w}^d) \textbf{\textsl{g}} \leq \bm{d},
\end{equation}
where~$\bm{M}$ is a signed selection matrix and~$\bm{d}$ is a vector of bounds that define the actuator limits and restrict the load sway to remain below a certain threshold.

The nonparametric optimal trajectory generation problem is  formulated as the following convex optimization problem:
\begin{subequations}\label{eq_opt_traj_search_final}
\begin{align}
\noalign{\vspace{2mm}\hrule\vspace{1mm}}
\minimize_{\textbf{\textsl{g}}}\quad
& \| \bm{w}|^{\mathcal{I}} - \bm{\mathcal{H}}_L(\bm{w}^d)|^{\mathcal{I}}\textbf{\textsl{g}}\|_\mathbf{W}^2
+ \lambda \|\textbf{\textsl{g}}\|_1 \label{eq_opt_traj_search_final_cost}\\
&\quad + \mu \|\bm{w}_\text{ref}-\bm{\mathcal{H}}_L(\bm{w}^d)\textbf{\textsl{g}}\|_\mathbf{R}^2
+ \sigma \| \bm{\mathcal{D}}\bm{\mathcal{H}}_L(\bm{w}^d)\textbf{\textsl{g}}\|^2 \notag\\[0.8ex]
\text{subject to}\quad
& \bm{\mathcal{H}}_L(\bm{w}^d)\big|^{\mathcal{J}}\textbf{\textsl{g}} =
\begin{bmatrix}
\bm{w}_\text{start}^\mathrm{T} \
\bm{w}_\text{target}^\mathrm{T}
\end{bmatrix}^{\mathrm{T}}
\label{eq_opt_traj_search_final_eq_const}\\
& \bm{M}\bm{\mathcal{H}}_L(\bm{w}^d)\textbf{\textsl{g}} \leq \bm{d}
\label{eq_opt_traj_search_final_ineq_const}\\
\noalign{\vspace{1mm}\hrule\vspace{-4mm}}\notag
\end{align}
\end{subequations}
In this problem, the hyperparameters~$\mu$ and~$\sigma$ must be tuned alongside~$\lambda$. 
The following section outlines the practical workflow followed in this work, including a systematic approach for tuning the hyperparameters.

\vspace{1em}
\begin{remark}
One might argue that including both norms of Eqs.~\eqref{eq_opt_approximation} and~\eqref{eq_opt_reference} to the cost function is redundant, since all elements of~$\bm{w}|^{\mathcal{I}}$ and~$\bm{\mathcal{H}}_L(\bm{w}^d)|^{\mathcal{I}}$ already appear in~$\bm{w}_\text{ref}$ and~$\bm{\mathcal{H}}_L(\bm{w}^d)$, respectively. 
While this redundancy holds in our specific case, we retain both terms to preserve generality in the formulation, when the reference deviates from the actual position.
\end{remark}

\begin{remark}
In the linear noise-free case, any solution~$\textbf{\textsl{g}}_p$ of Problem~\eqref{eq_opt_approximation} can be used to identify the space of trajectories that interpolate the start and target conditions,
\begin{equation}\label{eq_N_prime}
 \mathcal{G} := \textbf{\textsl{g}}_p + \text{null}\,(\bm{\mathcal{H}}_L(\bm{w}^d)|^{\mathcal{I}}),
\end{equation}
yielding the affine space of trajectories,
\begin{equation}
 \mathcal{S} := \bm{\mathcal{H}}_L(\bm{w}^d)\,\mathcal{G} 
 = \bm{w}_p + \bm{\mathcal{H}}_L(\bm{w}^d)\,\text{null}(\bm{\mathcal{H}}_L(\bm{w}^d)|^{\mathcal{I}}). \label{eq_nonunique_sol}
\end{equation}
The null space of the submatrix $\bm{\mathcal{H}}_L(\bm{w}^d)\big|^{\mathcal{I}}$ characterizes the degrees of freedom available for interpolating between the start and target conditions.
Let~$\bm{\mathcal{N}}$ be a basis matrix spanning~$\bm{\mathcal{H}}_L(\bm{w}^d)\,\text{null}(\bm{\mathcal{H}}_L(\bm{w}^d)|^{\mathcal{I}})$.  
Then, for any vector~$\bm{\beta} \in \mathbb{R}^{\text{nullity}(\bm{\mathcal{H}}_L(\bm{w}^d)|^{\mathcal{I}})}$, the trajectory
\begin{equation}
    \bm{\hat{w}} = \bm{w}_p + \bm{\mathcal{N}}\,\bm{\beta} \in \mathcal{S}.
\end{equation}
Hence without the regularization~\eqref{eq_opt_total_variation}, one can generate an optimal trajectory by solving
\begin{equation*}
\setlength{\arraycolsep}{0pt}
\begin{array}{cl}
\hline\\[-2mm]
\minimize_{\bm{\beta}} \quad
& \| \bm{w}_p + \bm{\mathcal{N}}\,\bm{\beta} - \bm{w}_{\text{ref}} \|^2 \\[5mm]
\mathrm{subject\;to} \quad
&~\eqref{eq_opt_traj_search_final_eq_const} \\[2mm]
&~\eqref{eq_opt_traj_search_final_ineq_const}. \\[2mm]
\hline
\end{array}
\end{equation*}
When the constraints~\eqref{eq_opt_traj_search_final_eq_const} and~\eqref{eq_opt_traj_search_final_ineq_const} are omitted, this problem admits a closed-form solution
\begin{equation}
 \bm{\beta}^\star = (\bm{\mathcal{N}}^\text{T}\bm{\mathcal{N}})^{-1}\bm{\mathcal{N}}^\text{T}(\bm{w}_\text{ref} - \bm{w}_p).
\end{equation}
We define this formulation as the \textit{indirect nonparametric optimal trajectory generation} problem.  
When the data~$\bm{w}^d$ is noisy, the Hankel matrix~$\bm{\mathcal{H}}_L(\bm{w}^d)|^{\mathcal{I}}$ has almost always a full row rank. Given that~${\mathcal{I}}$ contains only few start and final samples, $\bm{\mathcal{H}}_L(\bm{w}^d)|^{\mathcal{I}}$ has more columns than rows and its column rank is deficient. However, its null space lacks a behavioral interpretation, making this method impractical.

\end{remark}

\section{Practical Workflow for Optimal Trajectory Generation}\label{chap_practical_workflow}
Section~\ref{chap_trajGen} presented the tools necessary for capturing nonlinear dynamics from noisy data and generating optimal trajectories. 
This section provides a step-by-step guide on how to effectively capture the dynamics of a rotary crane and determine suitable hyperparameters for generating optimal trajectories that suppress load-sway and minimize time and energy.

To this end, the rotary crane is operated at a sampling frequency of~$20\,\text{Hz}$ using a piecewise constant input. 
The load angles~$\theta_1$ and~$\theta_2$, as well as the boom angle~$\theta_4$ are assumed to be available as noisy measured outputs. 
The optimization problems are solved in MATLAB using the solver \texttt{MOSEK}~\cite{aps2019mosek}.
All main functions presented in this section, together with the generated data, are publicly available\footnote{Code and data: https://github.com/ikhemakhem/BehavioralTrajGenCrane}.

\subsection{Input Design and Data Generation}
As explained in Section~\ref{sec_nonlinear}, using the fundamental lemma, one can capture broad regions of nonlinear dynamics. 
Consequently, the input must be designed to target a specific operational region during data collection. 
The resulting nonparametric model provides an accurate representation of the dynamics within that region.
Additionally, the optimal trajectory emerges from linearly combining the data trajectories, as shown in Eq.~\eqref{eq_lemma_w}.
Consequently, the family of inputs used to generate the data directly influences the inputs generated by the optimal trajectory search.

In the crane case, we aim to capture the dynamics resulting from smooth inputs, as jerky trajectories induce high load sway. The boom velocity is computed using a sum of sines with randomly generated frequencies within a specified range,
\begin{equation}\label{eq_sum_of_sines}
 \dot{\theta}_4(t) = \sum_{i=1}^{n_\Sigma} \sin(f_i t), \quad f_i \sim \mathcal{N}(\xi, \tau^2),
\end{equation}
where~$n_\Sigma$ is the number of sines used, and~$\xi$ and~$\tau$ are the mean and standard deviation of a normal distribution.
This class of functions is in~$C^\infty$ (infinitely differentiable), guaranteeing the smoothness of the acceleration and jerk. 
Each generated velocity sequence is normalized to match the hardware limits and tapered at the start and end to ensure smooth acceleration from rest and smooth braking to rest.
\vspace{1em}
\begin{remark}
In the literature, as in~\cite{elokda_data-enabled_2021}, it is common to inject pseudorandom binary sequences to excite the system across a broad range in order to capture a wide spectrum of the dynamics.
However, pseudorandom inputs cause erratic movements, resulting in a dictionary of nonsmooth inputs and strong load oscillations. Such behavior is undesirable for optimal trajectory generation in rotary cranes.
In applications that tolerate sudden input changes, this may be a more suitable choice.

\end{remark}
\vspace{1em}

Since the system dynamics are unknown and the fundamental lemma, in its original form, is not directly applicable to noisy data, there is no clear criterion for determining when the collected data are sufficient. 
As a practical guideline, we recommend generating a sufficiently large dataset so that the resulting matrix has substantially more columns than rows.
In the crane case, and for the proposed class of input signals, we found that using approximately fifteen times more columns than rows provides reliable performance. 
Previous works have shown that, within the behavioral setting, relatively short data sequences are often sufficient even for nonlinear systems, when compared to other data-driven approaches~\cite{elokda_data-enabled_2021, markovsky_data-driven_2023}.

To reduce computational time when solving the optimization problem, the matrix size can be reduced by heuristically selecting the most informative~$\nu$ columns using a QR-decomposition~\cite{strang2019linear}. 
This decomposition reorders the columns according to their informational content.
Columns that contribute new directions to the column space are placed first, pushing redundant data to the end of the reordered matrix.

The resulting matrix is the further denoised by truncating all singular values below a threshold~$\delta$. 
Note that the singular value thresholding result in the loss of the Hankel structure. 
Depending on the level of noise present and the available knowledge about the dynamics, the singular value decomposition can be omitted or replaced with other denoising techniques.

The parameters~$\nu$ and~$\delta$, together with the weight of the sparse regularization~$\lambda$ are hyperparameters that need careful tuning. 
This can be automatically done in a grid search by quantifying the model’s ability to predict outputs from initial conditions and inputs.
In fact, when all inputs and the initial~$N_\text{ini}$-length sequence of a trajectory~$\bm{w}$ representing the initial conditions are known, solving the missing data recovery problem in Eq.~\eqref{eq_opt_lasso} determines the corresponding outputs using the dynamics embedded in~$\bm{\mathcal{H}}_L(\bm{w}^d)$. This process effectively performs a \emph{nonparametric simulation}.

\subsection{Grid Search for Nonparametric Simulation}\label{sec:Nonparametric_Simulation}
In a state-space of a dynamical system of lag~$\ell$, if~$N_\text{ini} > \ell$ initial samples of a trajectory~$\bm{w}$ are known, then the initial conditions are uniquely defined~\cite{markovsky_data-driven_2008}. 
If additionally all inputs in~$\bm{w}$ are known, then all outputs are uniquely defined.
Since, in a nonparametric framework, the system's lag~$\ell$ is unknown, a large enough empirical value of~$N_\text{ini}$ is chosen to ensure the uniqueness of the outputs. 
Let~$\bm{u}_\text{sim} \in \mathbb{R}^{m(L-N_\text{ini})}$ be the vector of all inputs in~$\bm{w}$ excluding the first~$N_\text{ini}$ samples.
To guarantee that the given initial conditions and input sequence do not deviate during the approximation in Eq.~\eqref{eq_opt_approximation}, an inequality constraint is added to the optimization problem~\eqref{eq_opt_lasso},
\begin{equation}\label{eq_constraint_simulation}
    \big\| \bm{P}\bm{\mathcal{H}}_L(\bm{w}^d)|^{\mathcal{I}}\textbf{\textsl{g}} - \begin{bmatrix}
        \bm{w}|^{\{1,\hdots,qN_\text{ini}\}} \\
        \bm{u}_\text{sim}
    \end{bmatrix} \big\|_\infty < \epsilon,
\end{equation}
where~$\bm{P}$ is a selection and permutation matrix,~$\epsilon$ a small positive constant, and~$\|\cdot\|_\infty$ the infinity norm.

By generating~$N_\text{test}$ trajectories~$\bm{w}^\text{test}_i$ of length~$L$ and hiding all outputs except the first~$N_\text{ini}$ samples, Problem~\eqref{eq_opt_lasso} combined with constraint~\eqref{eq_constraint_simulation} performs a nonparametric simulation to recover these outputs. 
We select the optimal tuple \((\delta^\star, \lambda^\star, \nu^\star)\) from the grid that minimize the objective
\begin{equation}\label{eq_opt_test}
    \sum_{i=1}^{N_\text{test}} \|\bm{w}^\text{test}_i-\hat{\bm{w}}^\text{test}_i\|^2,
\end{equation}
where $\hat{\bm{w}}^\text{test}_i$ are the recovered trajectories. 
Since most of the known elements correspond to the input components of the trajectory, the weight matrix~$\bm{W}$ in Problem~\eqref{eq_opt_lasso} is chosen as the identity, giving equal preference to all known elements. 
For the short trajectories considered in this work, a more refined tuning that differentiates the magnitudes of the initial outputs (angles) from those of the inputs (angular velocities) is not required.

\subsection{Simulative Study}\label{sec:Nonparametric_Simulation_simStudy}
In this section we present a simulative study to highlight the effectiveness of the above mentioned workflow in capturing nonlinear dynamics from noisy data.
To do so, we first introduce a state-space model of a rotary crane.
This is the only section in which a simulation model is used. All results beyond this section refer to the experimental crane.

\subsubsection{Dynamics of a Rotary Crane}
\label{sec_system_dynamics}
We use the rotary crane dynamics presented in~\cite{ouyang_anti-sway_2010} and simplify it by setting  
\begin{equation}
 \ddot{\theta}_3 \equiv \dot{\theta}_3 \equiv \ddot{l} \equiv \dot{l} \equiv 0.
\end{equation} 
The system has a single input~$u$ corresponding to the slewing angular acceleration,
\begin{equation}
 u = \ddot{\theta}_4.
\end{equation} 
We define the state vector~$\bm{x}$ as
\begin{equation}
 \bm{x} = \begin{bmatrix}
\theta_1 &
\theta_2 & 
\theta_4 &
\dot{\theta}_1 &
\dot{\theta}_2 &
\dot{\theta}_4
\end{bmatrix}^\mathrm{T},
\end{equation}
and assume that the load angles ($\theta_1$ and~$\theta_2$), as well as the boom slewing angle~$\theta_4$ and angular velocity~$\dot{\theta}_4$ are system outputs, which are artificially corrupted with random noise.
 
A key distinction from the experimental crane is that, while the experimental setup uses the boom angular velocity as the input, a state-space representation admits only acceleration as an input. 
Considering the boom angular velocity as an output and introducing the boom angular acceleration as the input instead adds one additional element per sample in the Hankel matrix, increasing~$q$ to~$q = 5$. 
This modification does not introduce any structural changes to the proposed method, since it does not explicitly distinguish between inputs and outputs within a trajectory.

Following~\cite{ouyang_anti-sway_2010}, the load and boom dynamics are given by,
\begin{subequations}\label{eq_dynamics_simplified}
\begin{align}
&\ddot{\theta}_1 = -\alpha_1^2 \theta_1 + \alpha_2\dot{\theta}_4^2 + 2\dot{\theta}_2\dot{\theta}_4 + \epsilon_1^{nl} \label{eq_dynamics_theta1_simplified} \\ 
&\ddot{\theta}_2 = -\alpha_1^2 \theta_2 - \alpha_2\ddot{\theta}_4 + \epsilon_2^{nl} \label{eq_dynamics_theta2_simplified} \\ 
&\ddot{\theta}_4 = u, \label{eq_dynamics_theta4_simplified}
\end{align}
\end{subequations}
with~$\alpha_1 = \sqrt{\nicefrac{g}{l}}$ and~$\alpha_2 = \nicefrac{(l_b \sin{\theta_3})}{l}$.
The terms~$\epsilon_1^{nl}$ and~$\epsilon_2^{nl}$ account for negligible nonlinear effects, omitted for clarity.

The crane dynamics provide a good example to study the proposed method. 
The boom angle~$\theta_4$ and its angular velocity~$\dot{\theta}_4$ evolve linearly with the input~$\ddot{\theta}_4$ and are decoupled from all other states. In contrast, the dynamics of the radial load sway~$\theta_1$ have significant nonlinear effects. 
Nonlinearities are less important for the tangential load sway angle~$\theta_2$ (see Eq.~\eqref{eq_dynamics_simplified}). 
In this example, trajectories of~$L = 300$ samples are studied, and the outputs are sampled at a frequency of~$20~\text{Hz}$ from a simulation of the dynamics, using the solver \texttt{ode45} in MATLAB. 
\ref{sec_controllability_observability} presents a controllability analysis of the considered system. While this analysis is model-based, it provides a useful tool for interpreting the results presented in the remainder of this work.

\subsubsection{Results of the Simulative Study}
The performance of the resulting data-driven model is evaluated by performing multiple nonparametric simulations and comparing them with the results from the model-based simulations. 
Ten different test trajectories are used to find the optimal hyperparameters~$\delta$,~$\nu$, and~$\lambda$, using the objective function in Eq.~\eqref{eq_opt_test}.
The grid search yields the hyperparameters~$\lambda^\star = 1.833\times10^{-5}$,~$\delta^\star = 0.0081$ and~$\nu^\star = 2000$.

We studied two different cases.  
In case~$1$, the system was simulated for about one minute, and the data was organized in a Hankel structure. The data was kept noise-free, and no column selection was performed, keeping the Hankel matrix intact, with~$\text{rank}(\bm{\mathcal{H}}_L(\bm{w}^d)) = 320$, which is slightly higher than~$mL + n = 306$. We then run a nonparametric simulation with~$\delta = 0$ and~$\lambda = \lambda^\star$.  
In case~$2$, the data was noisy, and the optimal parameters from the grid search were used in the nonparametric simulation. The system was excited for about 20 minutes, and the~$\nu^\star$ most important columns were kept for~$\bm{\mathcal{H}}_L(\bm{w}^d)$. The resulting matrix has full rank and contains more columns than rows.

\begin{figure}
\centering
\includegraphics{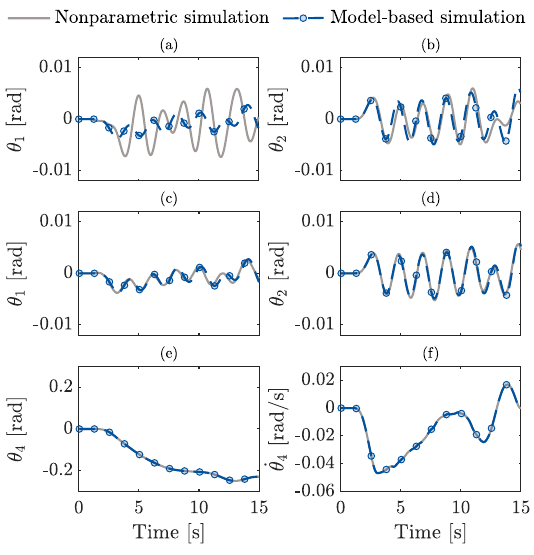}
\caption{Predicted outputs from the model-based and nonparametric simulations. Plots (a), (b), (e), and (f) show the results for case~1 (short noise-free data sequence, without QR-based column selection). Plots (c) and (d) highlight the improvement in predicting the load angles for case~2 (longer noisy data sequence with optimized hyperparameters), obtained by increasing the data length and tuning the hyperparameters.}
\label{fig_result_simulation}
\end{figure}

Fig.~\ref{fig_result_simulation} compares the results of both cases. The results from case~1 demonstrate that the model successfully captures the dynamics of~$\theta_4$ and~$\dot{\theta}_4$, as shown in Fig.~\ref{fig_result_simulation}(e) and~(f). 
This outcome aligns perfectly with the generalized fundamental lemma presented in Section~\ref{chap_fundamentals}, since the boom dynamics correspond to a linear double integrator, as described in Eq.~\eqref{eq_dynamics_theta4_simplified}.

The small Hankel matrix in case~1 also captures most of the dynamics of~$\theta_2$, as shown in Fig.~\ref{fig_result_simulation}(b). 
However, it fails to reproduce the behavior of~$\theta_1$, shown in Fig.~\ref{fig_result_simulation}(a). 
This limitation can be explained by the simplified load dynamics in Eq.~\eqref{eq_dynamics_simplified}. 
The dynamics of~$\theta_1$ exhibit significant nonlinear effects, and, as stated in~\ref{sec_controllability_observability}, linearizing the system leads to a loss of controllability.

Fig.~\ref{fig_result_simulation}(c) and~(d) in Fig.~\ref{fig_result_simulation} highlight the improvements obtained in capturing the dynamics of~$\theta_1$ and~$\theta_2$ in case~2. Unlike case~1, the larger Hankel matrix successfully accounts for the nonlinearities in the dynamics of~$\theta_1$, and a noticeable improvement is also observed in the predicted trajectory of~$\theta_2$.

\subsection{Nonparametric Model of the Experimental Crane} \label{sec_model_validation}
We generate a nonparametric model for the experimental rotary crane from data recorded using multiple input sequences from Eq.~\eqref{eq_sum_of_sines}. 
The dataset consists of trajectories totaling approximately 50 minutes. 
The data is organized in a Hankel matrix of depth~$L=500$ representing~$25$ second-long trajectories. 
Recall that the input in the experimental crane is the slewing angular velocity~$\dot{\theta}_4$.

The hyperparameters~$\lambda$,~$\delta$, and~$\nu$ were tuned using test trajectories separate from the modeling dataset. A grid search yielded the optimal values~$\nu^\star = 7000$,~$\delta^\star = 5 \times 10^{-6}$, and~$\lambda^\star = 10^{-8}$. 
The need for a significantly larger Hankel matrix compared to the simulative study in the previous section suggests the real crane measurements capture more nonlinear effects than those represented in Eq.~\eqref{eq_dynamics_simplified}.

\begin{figure}
    \centering
    \includegraphics{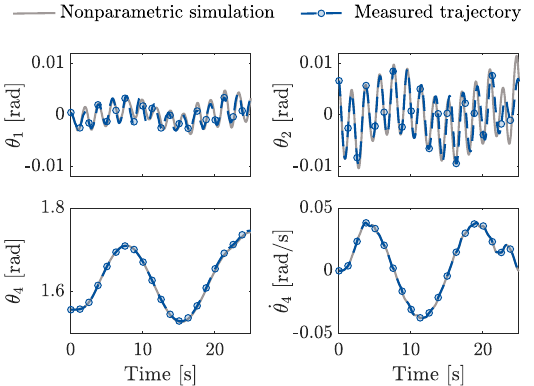} 
    \caption{Experimental measurements compared with the trajectory predicted by the nonparametric model under the same input and initial conditions.}
    \label{fig_model_validation_experiment}
\end{figure}

A comparison between the real data and the model's prediction is shown in Fig.~\ref{fig_model_validation_experiment}. 
The results demonstrate a good match between the two time series, with a slight divergence in the load angles towards the end of the trajectory.

\subsection{Hyperparameter Tuning for Trajectory Generation} \label{sec_hyperparameter_optimization_traj_gen}
Following the validation of the nonparametric model, optimal trajectories are generated directly by solving Eq.~\eqref{eq_opt_traj_search_final}. 
We set the weight matrix~$\mathbf{R}$ to identity and tune the weights~$\mu$,~$\sigma$, and~$\lambda$ using a grid search.
Note that the parameters~$\nu$ and~$\delta$ identified using the nonparametric simulation reflect inherent properties of the nonparametric model and are kept constant across all applications of the missing data recovery problem. 
However, the weight~$\lambda$ must be tuned again, since the additional regularization terms weighted by~$\mu$ and~$\sigma$ alters the problem structure, making ~$\lambda$,~$\mu$, and~$\sigma$  interdependent.

Finding suitable hyperparameters requires quantifying the quality of the generated trajectories. We evaluate the time to target, defined as when the boom stays within~$0.035\,\text{rad}$ of the target for at least~$5$ seconds, the load sway via the maximum and average of angles~$\theta_1$ and~$\theta_2$, the trajectory smoothness via the maximum and average central differences of the zero-padded velocity~$\dot{\theta}_4$, and the overshoot by integrating~$\theta_4$ when it exceeds the target.

Finding suitable hyperparameters requires quantifying the quality of the generated trajectories.  
We evaluate the time to target, defined as when the boom stays within~$0.035\,\text{rad}$ of the target for at least~$5$ seconds, the load sway via the maximum and average of angles~$\theta_1$ and~$\theta_2$, the trajectory smoothness via the maximum and average central differences of the zero-padded velocity~$\dot{\theta}_4$, and the overshoot by integrating~$\theta_4$ when it exceeds the target.

Each metric is normalized using empirical maximums and combined into a single weighted objective. To quantify feasibility, the trajectory input is applied to the system's state-space model, and a weighted 2-norm error from the optimal trajectory is added to the cost. This term is optional if no state-space model exists. For crane systems, simplified state-space models are generally available, allowing the addition of this term into the objective function to accelerate the tuning process.

\begin{figure}
\centering
\includegraphics{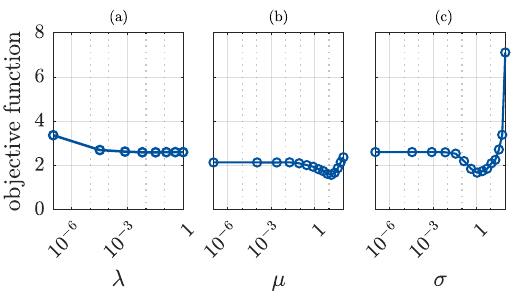} 
\caption{Changes in the objective function as one hyperparameter varies, 
with the other two fixed at their optimal values ($\lambda^\star = 0.0064$,~$\mu^\star = 14.3214$, and~$\sigma^\star = 2.5877$):
(a) Variation of~$\lambda$, (b) Variation of~$\mu$, (c) Variation of~$\sigma$.}
\label{fig_hyperparameter_optimization_traj_gen}
\end{figure}

Fig.~\ref{fig_hyperparameter_optimization_traj_gen} shows the objective function evolution for each hyperparameter while the others are held at optimal values. Both~$\mu$ and~$\sigma$ exhibit a plateau at low values, indicating minimal influence on the optimization cost, followed by a distinct minimum. The recorded optimal values are~$\lambda^\star = 0.0064$,~$\mu^\star = 14.3214$, and~$\sigma^\star = 2.5877$. 
Finding these weights was iterative but required significantly less effort than manual tuning.
The following section validates the resulting trajectory.

%
%
%
%

%
%
%
%
%
%
%

%
\section{Validation}\label{chap_experimental}
We evaluate the proposed method on the laboratory crane presented in Section~\ref{chap_setup}. An example trajectory is generated, tested, analyzed and then compared against the results obtained with a model-based approach.

\subsection{Data-Driven Optimal Trajectory} \label{sec_results_dd_traj}
\begin{figure}
    \centering
    \includegraphics{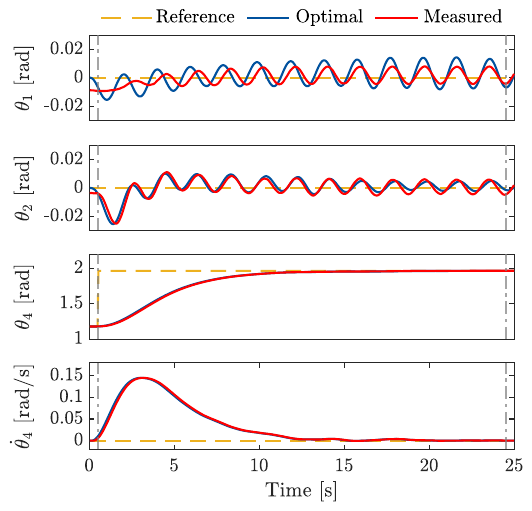} 
    \caption{Optimal trajectory generation results. The predicted optimal trajectory (input and output) is compared with experimental measurements. The number of known samples is~$N_\text{given}=10$ (time horizon~$0.5\,\text{s}$). The motion starts at~$\theta_4^\text{start}=\frac{3\pi}{8}\,\text{rad}$ and ends at~$\theta_4^\text{target}=\frac{5\pi}{8}\,\text{rad}$, with zero boom velocity and load sway at both endpoints. Gray vertical lines separate the interpolated region from the given initial and target data, and the orange curve denotes the reference trajectory~$\bm{w}_\text{ref}$.}
    \label{fig_results_DD}
\end{figure}
The number of known samples is set to~$N_\text{given} = 10$, corresponding to a~$0.5\,\text{s}$ time horizon.
The start and target boom angles are defined as~$\theta_4^\text{start} = \frac{3\pi}{8}\,\text{rad}$ and~$\theta_4^\text{target} = \frac{5\pi}{8}\,\text{rad}$, respectively, with zero boom velocity and load sway at the start and end of the trajectory.

Solving the optimization problem in Eq.~\eqref{eq_opt_traj_search_final} with the optimal hyperparameters yields the coefficient vector~$\textbf{\textsl{g}}_\text{opt}$, from which the optimal trajectory~$\bm{w}_\text{opt}$ is reconstructed via Eq.~\eqref{eq_lemma_w}. 
The corresponding optimal input sequence (boom angular velocity) is applied on the experimental crane, and the resulting boom position and load sway are recorded.

Fig.~\ref{fig_results_DD} compares the predicted optimal trajectory with the measured trajectory on the experimental crane. 
The gray vertical lines distinguish the middle interpolated region from the given initial and target sequences. 
The orange line defines the reference trajectory~$\bm{w}_\text{ref}$.

The optimal trajectory predicts that the boom reaches its target within~$15$ seconds, while the load sway remains below~$0.02\,\text{rad}$, except for a short initial swing in the tangential direction ($\theta_2$), where it slightly exceeds that bound. 
The experimental trajectory closely follows these predictions, reproducing the same transient swing observed in the early phase of the tangential motion. 
The control input effectively damps this initial disturbance, driving the sway smoothly to zero. 
In the radial direction, the measured sway remains even smaller than predicted, not exceeding~$0.01\,\text{rad}$. 
Although a slight increase in sway is visible toward the end of the motion, no significant oscillations are recorded.
The mismatch at the beginning is attributed to the experiment starting with the load slightly swaying in the radial direction.

Despite being executed in open-loop, the crane reaches the target with an error of~$0.002\,\text{rad}$. 
This high precision highlights the capacity of the data-driven model to account for hardware imperfections and to accurately capture the linear dynamics of the boom. 
The close match between the predicted and measured boom trajectories further supports the model’s fidelity.

It should be noted that achieving perfectly matching initial conditions between the optimal and experimental trajectories is practically infeasible, particularly due to the difficulty of setting the load exactly at rest. 
The overall agreement between the output predicted by the optimal trajectory generation problem and the experimental results demonstrates the validity of the underlying data-driven model and the accuracy of the generated trajectory.

\subsection{Comparison with a Model-Based Approach}\label{chap_comaprison}
To validate the performance of the proposed framework, we compare the generated trajectory with a model-based time-optimal trajectory that was optimized to fulfill the same requirements using a conventional approach.
As a benchmark, we use a multi-waypoint trajectory generation method that has proven successful in previous applications, in which smooth trajectories for tower cranes while suppressing both load sway and load rotation have been generated~\cite{farrage2025optimal, burkhardt2023optimization}.   

\subsubsection{Model-Based Optimal Trajectory Generation}

Conventional trajectory generation methods typically pursue similar objectives, namely minimizing motion time and suppressing load sway.
However, their optimization structures have been limited with respect to trajectory profiles and the acceptable values of the swaying angles.
The proposed model-based optimization overcomes these drawbacks by enabling flexible trajectory generation with reduced computational effort~\cite{farrage2025optimal}.

The boom slewing trajectory is discretized into \(N_{\mathrm{wp}}=16\) waypoints, resulting in~\(15\) segments of equal duration~\(\tau\).
The decision variables are the boom angle \(\theta_4(k)\), the corresponding angular velocity~\(\dot{\theta}_4(k)\), the angular acceleration~\(\ddot{\theta}_4(k)\) at the waypoints, ($k=1,\ldots,N_{\mathrm{wp}}$), and the segment duration~\(\tau\).
The optimization problem minimizes a weighted sum of the segment duration and the deviation from a prescribed reference boom trajectory \(\theta_4^{\mathrm{ref}}(k)\), which is defined as a linear interpolation between the start and target boom angles,
\begin{equation*}\label{eq_model_based}
\setlength{\arraycolsep}{0pt}
\begin{array}{cl}
\hline\\[-2mm]
\minimize_{\substack{\theta_4(1),\ldots,\theta_4(N_{\mathrm{wp}}),\\
\dot{\theta}_4(1),\ldots,\dot{\theta}_4(N_{\mathrm{wp}}),\\
\ddot{\theta}_4(1),\ldots,\ddot{\theta}_4(N_{\mathrm{wp}}),\,T}}
&
\tau
+ \dfrac{\sigma^2}{ N_{\mathrm{wp}}}
\displaystyle\sum_{k=1}^{N_{\mathrm{wp}}}
\bigl(\theta_4(k)-\theta_4^{\mathrm{ref}}(k)\bigr)^2
\\[7mm]
\mathrm{subject\;to}
& \theta_4(1)=\theta_4^{\mathrm{ref}}(1), \quad
  \theta_4(N_{\mathrm{wp}})=\theta_4^{\mathrm{ref}}(N_{\mathrm{wp}}), \\[3pt]
& \dot{\theta}_4(1)=0, \qquad\quad
  \dot{\theta}_4(N_{\mathrm{wp}})=0, \\[3pt]
& \ddot{\theta}_4(1)=0, \qquad\quad
  \ddot{\theta}_4(N_{\mathrm{wp}})=0, \\[4pt]
& \dot{\theta}_4(k+1)
= \frac{\theta_4(k)-\theta_4(k-1)}{\tau}, \\[4pt]
& \ddot{\theta}_4(k+1)
= \frac{\dot{\theta}_4(k)-\dot{\theta}_4(k-1)}{\tau}, \\[4pt]
& |\dot{\theta}_4(k)| \le \dot{\theta}_4^{\max}, \quad\;
  |\ddot{\theta}_4(k)| \le \ddot{\theta}_4^{\max}, \\[4pt]
& |\theta_p(t_k)| \le \theta_{\mathrm{sw}}, \quad\;
  |\dot{\theta}_p(t_k)| \le \dot{\theta}_{\mathrm{sw}}, \\[3pt]
& |\theta_p(t_f)| \le \theta^{f}, \quad\;\:
  |\dot{\theta}_p(t_f)| \le \dot{\theta}^{f}, \\[4mm]
\hline
\end{array}
\end{equation*}
where $p\in\{1,2\}$,~$k =1,...,N_\mathrm{wp}-1$, and \(t_f=(N_{\mathrm{wp}}-1)\tau\), and the weighting parameter is set to \(\sigma=10~\mathrm{rad}^{-1}\).

Kinematic consistency between the discretized boom angle, angular velocity and angular acceleration profiles is enforced through the finite-difference equality constraints.
The load sway is constrained at all discretization points along the trajectory by bounds on the sway angles (\(\theta_1\), and \(\theta_2\)) and angular velocities (\(\dot{\theta}_1\), and \(\dot{\theta}_2\)).
Tighter bounds are imposed at the final segment to ensure near-zero residual oscillations.
The sway states are obtained by numerically integrating the load dynamic model over each segment using piecewise-constant boom angular velocity and acceleration.

The numerical values of the bounds used in the optimization are summarized in Table~\ref{tab:constraints}.
The resulting nonlinear program is solved using a sequential quadratic programming method implemented in the \texttt{fmincon} function in MATLAB.

\begin{table}[t]
\centering
\setlength{\tabcolsep}{12pt}
\begin{tabular}{lcc}
\hline
Parameter & Units & Value \\
\hline
$\ddot{\theta}_{4}^{\max}$    & [rad/s$^{2}$]    & 0.01724 \\
$\dot{\theta}_{4}^{\max}$     & [rad/s]          & 0.1724 \\
$\theta_{\mathrm{sw}}$        & [rad]            & 0.0349 \\
$\dot{\theta}_{\mathrm{sw}}$ & [rad/s]          & 0.0175 \\
$\theta^{f}$                  & [rad]            & 0.0017 \\
$\dot{\theta}^{f}$            & [rad/s]          & 0.0087 \\
\hline
\end{tabular}
\caption{Constraint values used in the model-based optimization. \label{tab:constraints}}
\end{table}

\subsubsection{Comparison}

\begin{figure}
\centering
\includegraphics{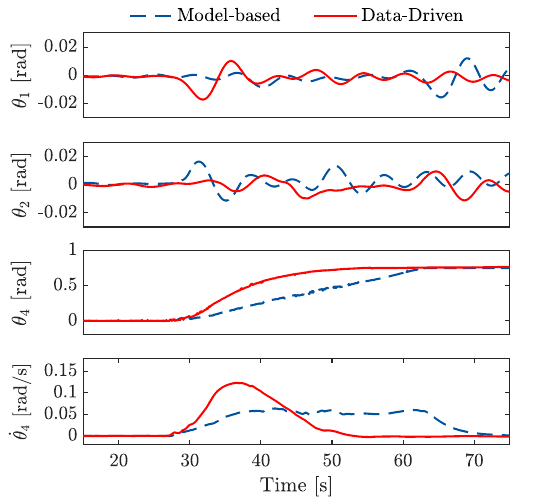}
\caption{Comparison of the trajectories generated by the proposed data-driven method and a benchmark model-based approach.}
\label{fig_comparison_modelbased}
\end{figure}
Both the data-driven and model-based methods were tested under identical conditions. 
Similar to the experiment in Section~\ref{sec_results_dd_traj}, the optimal motion rotates the boom by one eighth of a full turn, this time from~$\theta_4^\text{start} = 0$ to~$\theta_4^\text{target} = \frac{\pi}{4}\,\text{rad}$. 
To assess the repeatability of the results, each experiment was conducted five times for both trajectory generation methods. 

Fig.~\ref{fig_comparison_modelbased} compares the optimal trajectories obtained from the data-driven and model-based approaches. 
The data-driven method yielded faster and smoother motion profiles. 
The model-based approach missed the target by~$0.0315\,\text{rad}$, corresponding to a~$75\,\%$ larger error than the behavioral method.
It required approximately \(35\) seconds to reach the target, which is more than twice the duration of the data-driven trajectory.
The proposed approach achieved smaller sway amplitudes than the model-based method, reducing the swaying angles~$\theta_1$ and~$\theta_2$ by~$9.2\%$ and~$37.95\%$, respectively.

The proposed behavioral method resulted in faster motion, reduced load sway, and lower tracking error compared to the state-of-the-art model-based benchmark.
This performance gain is mainly due to differences in the followed reference trajectories. 
The data-driven method considers an abrupt jump in the slewing angle with zero velocity and acceleration, whereas the benchmark relies on a linear transition between waypoints, which inherently limits the achievable time performance. In the benchmark method, as in most classical model-based trajectory generation approaches, such jump references typically induce strong transients and increased load oscillations.
Despite operating under this more demanding reference, the data-driven method generates smooth and fast input trajectories. This can be attributed to the fact that the optimization is performed over a dictionary of measured input–output data, ensuring dynamical consistency with the true crane behavior, as well as to the convex formulation of the trajectory generation problem. It should also be noted that the cost function of the data-driven approach explicitly accounts for energetic efficiency, providing an additional advantage over the benchmark method.

\section{Conclusions}\label{chap_conclusion}
This study proposed a novel trajectory optimization framework for open-loop crane control. In particular, it addressed the reduction of load sway during the slewing motion of rotary cranes. The framework designs slewing trajectories that jointly optimize load sway, travel time, and energy consumption to move a load from an initial to a target position. 
In doing so, a highly accurate nonparametric model of the restricted crane behavior was identified, enabling the generation of faster trajectories with reduced load sway and tracking error compared to a parametric baseline.

With this work, we reduce the complexity of trajectory optimization for a nonlinear underactuated and nonlinearly accessible system to linear algebraic operations and convex optimization problems. 
We show that crane automation can be achieved without relying on expert on-site operators, detailed system modeling, high-fidelity simulators, or complex nonlinear solvers. 
Although our approach is data-driven, the results indicate that it is not data-greedy compared to common data-driven methods.

Nevertheless, the proposed method requires careful tuning of multiple hyperparameters. 
Additionally, both the identified nonparametric model and the generated trajectories strongly depend on the input design. This dependence may become critical in more complex applications, where designing sufficiently informative inputs becomes challenging and requires additional effort.

While the proposed method builds upon foundational results in systems theory, specifically Willems' fundamental lemma, a rigorous theoretical analysis of the behavioral framework for nonlinear dynamics remains an open problem and should be addressed in future work.
Further research should also validate the approach on a full-scale crane and extend it to account for unforeseen disturbances such as wind and torsional effects.  
In this context, a natural extension to this work would be to apply Data-enabled Predictive Control (DeePC), presented in~\cite{coulson_data-enabled_2019}, for closed-loop optimal control.
Moreover, evaluating the proposed method on other crane types would further assess its generalization capabilities.

\appendix
    
\section{Controllability Analysis} \label{sec_controllability_observability}
The rotary crane is controlled only by its slewing motion and hence is underactuated. 
With a single control input, the goal is to control three degrees of freedom. 
To establish the foundation for anti-sway motion planning for such a system, one must first ensure that the system is controllable.
We use the rotary crane's dynamics presented in \ref{sec_system_dynamics}.

The system is control-affine.
Its state-space representation is given by,
\begin{align}\label{eq_state_space_representation}
\dot{\bm{x}} &= \bm{F}(\bm{x}) + \bm{b}u \\
\bm{y} &= \bm{C}\bm{x}.
\end{align}
Linearizing the system around the downward equilibrium at any boom position results in a loss of controllability for the linearized model. However, the system remains nonlinearly reachable, meaning its control input can generate motion along the non-actuated directions.

This can be verified using the Lie Algebra Rank Condition~\cite{sussmann_controllability_1972}. 
Consider the vector fields~$\bm{F}$ and~$\bm{b}$, the Lie Bracket between~$\bm{F}$ and~$\bm{b}$ is given by
\begin{equation}
 [\bm{F}, \bm{b}] = \frac{\partial{\bm{b}}}{\partial{\bm{x}}} \bm{F} - \frac{\partial{\bm{F}}}{\partial{\bm{x}}} \bm{b}.
\end{equation}
Any higher-order Lie bracket can be computed recursively by replacing~$\bm{b}$ with the previously computed Lie bracket.

For an arbitrary point in the state space~$\bar{\bm{x}}$, 
the controllability matrix~$\bm{\mathcal{Q}}$ is given by
\begin{equation}
\bm{\mathcal{Q}}(\bar{\bm{x}}) =
\begin{bmatrix}
0 & 0 & 2\bar{\dot{\theta}}_2 & -4\alpha_1^2\bar{\theta}_2 & -2\alpha_1^2 d_1 & 16\alpha_1^4\bar{\theta}_2 \\
0 & -2\bar{\dot{\theta}}_2 & 2\alpha_1^2\bar{\theta}_2 & 2\alpha_1^2 d_2 & -8\alpha_1^4\bar{\theta}_2 & -4\alpha_1^4 d_1 \\
0 & \alpha_2 & 0 & -\alpha_1^2\alpha_2 & 0 & \alpha_1^4\alpha_2 \\
-\alpha_2 & 0 & \alpha_1^2\alpha_2 & 0 & -\alpha_1^4\alpha_2 & 0 \\
0 & -1 & 0 & 0 & 0 & 0 \\
1 & 0 & 0 & 0 & 0 & 0
\end{bmatrix}.
\end{equation}
with~$d_1 = (4\bar{\dot{\theta}}_2 - \alpha_2 \bar{\dot{\theta}}_4)$ and~$d_2 = (2\bar{\dot{\theta}}_2 - \alpha_2 \bar{\dot{\theta}}_4)$.
Each column of~$\bm{\mathcal{Q}}$ represents a Lie bracket of the vector fields~$\bm{F}$ and~$\bm{b}$, from order 0 to order 5.
The determinant of~$\bm{\mathcal{Q}}$ is,
\begin{equation}
\mathrm{det}(\bm{\mathcal{Q}}) = 4 \alpha_1^{10} \alpha_2^{2}
\left(18 \alpha_1^{2} \bar{\theta}_2^{2} + \alpha_2^{2} \bar{\dot{\theta}}_4^{2} - 9 \alpha_2 \bar{\dot{\theta}}_2 \bar{\dot{\theta}}_4 + 18 \bar{\dot{\theta}}_2^{2}
\right).
\end{equation}

The zero set of~$\det(\bm{\mathcal{Q}})$ has Lebesgue measure zero in the~$(\bar{\theta}_2,\bar{\dot{\theta}}_2,\bar{\dot{\theta}}_4)$-space.
Hence,~$\det(\bm{\mathcal{Q}})\neq 0$ for almost all points in this domain.
Consequently, the system is nonlinearly accessible almost everywhere in the considered operating region.

Motion planning for nonlinear underactuated systems that are only nonlinearly accessible requires exploiting the nonlinear control couplings revealed by the Lie brackets computed in the controllability matrix. 
These controls are often non-intuitive, rendering the control of such systems by a human operator a challenging task. 
Additionally, being linearly uncontrollable renders all classical trajectory planning methods based on linearization, such as LQR and linear MPC, infeasible. 
One common method for planning trajectories for such systems is the use of differential geometry, which exploits the \textit{motion primitives} explored through the Lie brackets to generate a trajectory.
This work follows the same idea of using motion primitives to generate trajectories for underactuated, linearly uncontrollable systems, yet it proposes a method that succeeds in leveraging linearization properties to produce optimal trajectories.

\section*{Acknowledgments}
The experimental system for this work was provided by Kobelco Construction Machinery Co., Ltd. 
We would like to sincerely thank them for providing the system. 
Iskandar Khemakhem was supported by the DAAD PROMOS program and the International Max Planck
Research School for Intelligent Systems (IMPRS-IS). 

\section*{CRediT authorship contribution statement}
\textbf{Iskandar Khemakhem:} Conceptualization, Methodology, Software, Validation, Formal analysis, Investigation, Data curation, Writing – original draft, review~$\&$ editing, Visualization. \textbf{Manuel Zobel:} Formal analysis, Writing – review~$\&$ editing. \textbf{Johannes Sch{\"u}le:} Supervision, Writing – review~$\&$ editing. \textbf{Oliver Sawodny:} Supervision, Resources. \textbf{Naoki Uchiyama:} Supervision, Writing – review~$\&$ editing, Resources. \textbf{Abdallah Farrage:} Methodology, Validation, Writing – review~$\&$ editing.

\section*{Declaration of competing interest}
All authors declare no competing interest.

\section*{Data availability}
The data supporting the findings of this study are available upon request.
\section*{Declaration of generative AI in the manuscript preparation process}
During the preparation of this work, the authors used ChatGPT (OpenAI) to assist with language refinement and minor text editing. The authors reviewed and edited all content as needed and take full responsibility for the content of the published article.

\bibliographystyle{elsarticle-num}
\bibliography{MyLibrary}

@article{hou_model-based_2013,
	title = {From model-based control to data-driven control: Survey, classification and perspective},
	volume = {235},
	issn = {0020-0255},
	url = {https://www.sciencedirect.com/science/article/pii/S0020025512004781},
	doi = {10.1016/j.ins.2012.07.014},
	series = {Data-based Control, Decision, Scheduling and Fault Diagnostics},
	shorttitle = {From model-based control to data-driven control},
	pages = {3--35},
	journaltitle = {Information Sciences},
	shortjournal = {Information Sciences},
	author = {Hou, Zhong-Sheng and Wang, Zhuo},
	urldate = {2024-02-19},
	date = {2013},
}

@article{dorfler_bridging_2023,
	title = {Bridging Direct and Indirect Data-Driven Control Formulations via Regularizations and Relaxations},
	volume = {68},
	rights = {https://ieeexplore.ieee.org/Xplorehelp/downloads/license-information/{IEEE}.html},
	issn = {0018-9286, 1558-2523, 2334-3303},
	url = {https://ieeexplore.ieee.org/document/9705109/},
	doi = {10.1109/TAC.2022.3148374},
	pages = {883--897},
	number = {2},
	journaltitle = {{IEEE} Transactions on Automatic Control},
	shortjournal = {{IEEE} Trans. Automat. Contr.},
	author = {Dorfler, Florian and Coulson, Jeremy and Markovsky, Ivan},
	urldate = {2024-05-07},
	date = {2023},
	langid = {english},
}

@article{markovsky_identifiability_2023,
	title = {Identifiability in the Behavioral Setting},
	volume = {68},
	rights = {https://ieeexplore.ieee.org/Xplorehelp/downloads/license-information/{IEEE}.html},
	issn = {0018-9286, 1558-2523, 2334-3303},
	url = {https://ieeexplore.ieee.org/document/9904308/},
	doi = {10.1109/TAC.2022.3209954},
	pages = {1667--1677},
	number = {3},
	journaltitle = {{IEEE} Transactions on Automatic Control},
	shortjournal = {{IEEE} Trans. Automat. Contr.},
	author = {Markovsky, Ivan and Dorfler, Florian},
	urldate = {2024-05-07},
	date = {2023},
	langid = {english},
}

@article{elokda_data-enabled_2021,
	title = {Data-enabled predictive control for quadcopters},
	volume = {31},
	rights = {© 2021 The Authors. International Journal of Robust and Nonlinear Control published by John Wiley \& Sons Ltd.},
	issn = {1099-1239},
	url = {https://onlinelibrary.wiley.com/doi/abs/10.1002/rnc.5686},
	doi = {10.1002/rnc.5686},
	pages = {8916--8936},
	number = {18},
	journaltitle = {International Journal of Robust and Nonlinear Control},
	author = {Elokda, Ezzat and Coulson, Jeremy and Beuchat, Paul N. and Lygeros, John and Dörfler, Florian},
	urldate = {2024-05-07},
	date = {2021},
	langid = {english},
}

@article{willems_note_2005,
title = {A note on persistency of excitation},
journal = {Systems \& Control Letters},
volume = {54},
number = {4},
pages = {325-329},
year = {2005},
issn = {0167-6911},
doi = {https://doi.org/10.1016/j.sysconle.2004.09.003},
url = {https://www.sciencedirect.com/science/article/pii/S0167691104001434},
author = {Jan C. Willems and Paolo Rapisarda and Ivan Markovsky and Bart L.M. {De Moor}}
}

@article{markovsky_data-driven_2023,
	title = {Data-Driven Control Based on the Behavioral Approach: From Theory to Applications in Power Systems},
	volume = {43},
	rights = {https://ieeexplore.ieee.org/Xplorehelp/downloads/license-information/{IEEE}.html},
	issn = {1066-033X, 1941-000X},
	url = {https://ieeexplore.ieee.org/document/10266847/},
	doi = {10.1109/MCS.2023.3291638},
	shorttitle = {Data-Driven Control Based on the Behavioral Approach},
	pages = {28--68},
	number = {5},
	journaltitle = {{IEEE} Control Systems},
	shortjournal = {{IEEE} Control Syst.},
	author = {Markovsky, Ivan and Huang, Linbin and Dörfler, Florian},
	urldate = {2024-05-07},
	date = {2023},
	langid = {english},
}

@article{markovsky_data-driven_2022,
	title = {Data-driven dynamic interpolation and approximation},
	volume = {135},
	doi = {10.1016/j.automatica.2021.110008},
	pages = {110008},
	journaltitle = {Automatica},
	shortjournal = {Automatica},
	author = {Markovsky, Ivan and Dörfler, Florian},
	date = {2022},
}

@inproceedings{coulson_data-enabled_2019,
	location = {Naples, Italy},
	title = {Data-Enabled Predictive Control: In the Shallows of the {DeePC}},
	url = {https://ieeexplore.ieee.org/document/8795639},
	doi = {10.23919/ECC.2019.8795639},
	shorttitle = {Data-Enabled Predictive Control},
	pages = {307--312},
	booktitle = {2019 18th European Control Conference ({ECC})},
	author = {Coulson, Jeremy and Lygeros, John and Dörfler, Florian},
	urldate = {2024-05-08},
	date = {2019},
}

@article{eckart_approximation_1936,
	title = {The approximation of one matrix by another of lower rank},
	volume = {1},
	issn = {1860-0980},
	url = {https://doi.org/10.1007/BF02288367},
	doi = {10.1007/BF02288367},
	pages = {211--218},
	number = {3},
	journaltitle = {Psychometrika},
	shortjournal = {Psychometrika},
	author = {Eckart, Carl and Young, Gale},
	urldate = {2024-05-27},
	date = {1936},
	langid = {english},
}

@article{markovsky_data-driven_2008,
	title = {Data-driven simulation and control},
	volume = {81},
	issn = {0020-7179},
	url = {https://doi.org/10.1080/00207170801942170},
	doi = {10.1080/00207170801942170},
	pages = {1946--1959},
	number = {12},
	journaltitle = {International Journal of Control},
	author = {Markovsky, Ivan and Rapisarda, Paolo},
	urldate = {2024-05-29},
	date = {2008},
}

@inproceedings{ouyang_anti-sway_2010,
	location = {Yokohama, Japan},
	title = {Anti-sway control of rotary crane only by horizontal boom motion},
	url = {https://ieeexplore.ieee.org/document/5611096/?arnumber=5611096},
	doi = {10.1109/CCA.2010.5611096},
	pages = {591--595},
	booktitle = {2010 {IEEE} International Conference on Control Applications},
	author = {Ouyang, Huimin and Uchiyama, Naoki and Sano, Shigenori},
	urldate = {2024-07-21},
	date = {2010},
	note = {{ISSN}: 1085-1992},
}

@article{sussmann_controllability_1972,
	title = {Controllability of nonlinear systems},
	volume = {12},
	issn = {0022-0396},
	url = {https://www.sciencedirect.com/science/article/pii/0022039672900071},
	doi = {10.1016/0022-0396(72)90007-1},
	pages = {95--116},
	number = {1},
	journaltitle = {Journal of Differential Equations},
	shortjournal = {Journal of Differential Equations},
	author = {Sussmann, Héctor J and Jurdjevic, Velimir},
	urldate = {2024-07-21},
	date = {1972},
}

@article{takahashi_sensor-less_2022,
	title = {Sensor-less and time-optimal control for load-sway and boom-twist suppression using boom horizontal motion of large cranes},
	volume = {134},
	issn = {0926-5805},
	url = {https://www.sciencedirect.com/science/article/pii/S0926580521005379},
	doi = {10.1016/j.autcon.2021.104086},
	pages = {104086},
	journaltitle = {Automation in Construction},
	shortjournal = {Automation in Construction},
	author = {Takahashi, Hideki and Farrage, Abdallah and Terauchi, Kenichi and Sasai, Shintaro and Sakurai, Hitoshi and Okubo, Masaki and Uchiyama, Naoki},
	urldate = {2024-07-31},
	date = {2022},
}

@inproceedings{regina_lauer_state_2023,
	location = {Atlanta, {GA}, {USA}},
	title = {State Estimation with Static Displacement Compensation for Large-Scale Manipulators},
	url = {https://ieeexplore.ieee.org/abstract/document/10039134},
	doi = {10.1109/SII55687.2023.10039134},
	pages = {1--6},
	booktitle = {2023 {IEEE}/{SICE} International Symposium on System Integration ({SII})},
	author = {Regina Lauer, Anja Patricia and Lerke, Otto and Gienger, Andreas and Schwieger, Volker and Sawodny, Oliver},
	urldate = {2024-08-14},
	date = {2023},
	note = {{ISSN}: 2474-2325},
}

@article{farrage_trajectory_2023,
	title = {Trajectory generation of rotary cranes based on A* algorithm and time-optimization for obstacle avoidance and load-sway suppression},
	volume = {94},
	issn = {0957-4158},
	url = {https://www.sciencedirect.com/science/article/pii/S0957415823000818},
	doi = {10.1016/j.mechatronics.2023.103025},
	pages = {103025},
	journaltitle = {Mechatronics},
	shortjournal = {Mechatronics},
	author = {Farrage, Abdallah and Takahashi, Hideki and Terauchi, Kenichi and Sasai, Shintaro and Sakurai, Hitoshi and Okubo, Masaki and Uchiyama, Naoki},
	urldate = {2024-08-14},
	date = {2023},
}

@article{berberich_data-driven_2021,
	title = {Data-Driven Model Predictive Control With Stability and Robustness Guarantees},
	volume = {66},
	issn = {1558-2523},
	url = {https://ieeexplore.ieee.org/document/9109670/?arnumber=9109670},
	doi = {10.1109/TAC.2020.3000182},
	pages = {1702--1717},
	number = {4},
	journaltitle = {{IEEE} Transactions on Automatic Control},
	author = {Berberich, Julian and Köhler, Johannes and Müller, Matthias A. and Allgöwer, Frank},
	urldate = {2024-08-14},
	date = {2021},
}

@article{carron_data-driven_2019,
	title = {Data-Driven Model Predictive Control for Trajectory Tracking With a Robotic Arm},
	volume = {4},
	issn = {2377-3766},
	url = {https://ieeexplore.ieee.org/abstract/document/8768048},
	doi = {10.1109/LRA.2019.2929987},
	pages = {3758--3765},
	number = {4},
	journaltitle = {{IEEE} Robotics and Automation Letters},
	author = {Carron, Andrea and Arcari, Elena and Wermelinger, Martin and Hewing, Lukas and Hutter, Marco and Zeilinger, Melanie N.},
	urldate = {2024-08-14},
	date = {2019},
}

@inproceedings{wolff_nonlinear_2022,
	location = {Jeju, Korea},
	title = {Nonlinear Model Predictive Control with Non-Equidistant Discretization Time Grids for Rotary Cranes},
	url = {https://ieeexplore.ieee.org/document/9828180/?arnumber=9828180},
	doi = {10.23919/ASCC56756.2022.9828180},
	pages = {1753--1758},
	booktitle = {2022 13th Asian Control Conference ({ASCC})},
	author = {Wolff, Frank and Uchiyama, Naoki and Burkhardt, Mark and Sawodny, Oliver},
	urldate = {2024-08-14},
	date = {2022},
	note = {{ISSN}: 2770-8373},
}

@inproceedings{arnold_trajectory_2007,
	location = {Singapore},
	title = {Trajectory Tracking for Boom Cranes Based on Nonlinear Control and Optimal Trajectory Generation},
	url = {https://ieeexplore.ieee.org/abstract/document/4389439},
	doi = {10.1109/CCA.2007.4389439},
	pages = {1444--1449},
	booktitle = {2007 {IEEE} International Conference on Control Applications},
	author = {Arnold, Eckhard and Neupert, Jorg and Sawodny, Oliver and Schneider, Klaus},
	urldate = {2024-08-14},
	date = {2007},
	note = {{ISSN}: 1085-1992},
}

@article{ouyang_s_curve_2011,
  title={S-curve trajectory generation for residual load sway suppression in a rotary crane system using only horizontal boom motion},
  author={Ouyang, Huimin and Uchiyama, Naoki and Sano, Shigenori},
  journal={Journal of System Design and Dynamics},
  volume={5},
  number={7},
  pages={1418--1432},
  year={2011},
  publisher={The Japan Society of Mechanical Engineers},
  doi={https://doi.org/10.1299/jsdd.5.1418}
}

@article{willems1986time,
title = {From time series to linear system—Part I. Finite dimensional linear time invariant systems},
journal = {Automatica},
volume = {22},
number = {5},
pages = {561-580},
year = {1986},
issn = {0005-1098},
doi = {https://doi.org/10.1016/0005-1098(86)90066-X},
url = {https://www.sciencedirect.com/science/article/pii/000510988690066X},
author = {Jan C. Willems}
}

@ARTICLE{van_waarde_extension,
  author={van Waarde, Henk J. and De Persis, Claudio and Camlibel, M. Kanat and Tesi, Pietro},
  journal={IEEE Control Systems Letters}, 
  title={Willems’ Fundamental Lemma for State-Space Systems and Its Extension to Multiple Datasets}, 
  year={2020},
  volume={4},
  number={3},
  pages={602-607},
  doi={10.1109/LCSYS.2020.2986991}}

@ARTICLE{Coulson_distributionally,
  author={Coulson, Jeremy and Lygeros, John and Dörfler, Florian},
  journal={IEEE Transactions on Automatic Control}, 
  title={Distributionally Robust Chance Constrained Data-Enabled Predictive Control}, 
  year={2022},
  volume={67},
  number={7},
  pages={3289-3304},
  doi={10.1109/TAC.2021.3097706}}

@book{strang2019linear,
  title={Linear algebra and learning from data},
  author={Strang, Gilbert and others},
  volume={4},
  year={2019},
  publisher={Wellesley-Cambridge Press Cambridge}
}

@article{proctor2018generalizing,
  title={Generalizing Koopman theory to allow for inputs and control},
  author={Proctor, Joshua L and Brunton, Steven L and Kutz, J Nathan},
  journal={SIAM Journal on Applied Dynamical Systems},
  volume={17},
  number={1},
  pages={909--930},
  year={2018},
  publisher={SIAM},
  doi={https://doi.org/10.1137/16M1062296}
}

@article{amini2025carleman,
  title={Carleman linearization of nonlinear systems and its finite-section approximations},
  author={Amini, Arash and Zheng, Cong and Sun, Qiyu and Motee, Nader},
  journal={Discrete and Continuous Dynamical Systems-B},
  volume={30},
  number={2},
  pages={577--603},
  year={2025},
  publisher={Discrete and Continuous Dynamical Systems-B}
}

@article{alsalti2023data,
  author={Alsalti, Mohammad and Lopez, Victor G. and Berberich, Julian and Allgöwer, Frank and Müller, Matthias A.},
  journal={IEEE Transactions on Automatic Control}, 
  title={Data-Based Control of Feedback Linearizable Systems}, 
  year={2023},
  volume={68},
  number={11},
  pages={7014-7021},
  doi={10.1109/TAC.2023.3249289}}

@article{yuan2022data,
  title={Data-driven optimal control of bilinear systems},
  author={Yuan, Zhenyi and Cort{\'e}s, Jorge},
  journal={IEEE Control Systems Letters},
  volume={6},
  pages={2479--2484},
  year={2022},
  publisher={IEEE},
  doi={10.1109/LCSYS.2022.3164983}
}

@inproceedings{berberich2020trajectory,
  title={A trajectory-based framework for data-driven system analysis and control},
  author={Berberich, Julian and Allg{\"o}wer, Frank},
  booktitle={2020 European Control Conference (ECC)},
  pages={1365--1370},
  year={2020},
  organization={IEEE},
  doi={10.23919/ECC51009.2020.9143608}
}

@misc{lazar2025product, 
  title={From Product Hilbert Spaces to the Generalized Koopman Operator and the Nonlinear Fundamental Lemma},
  author={Lazar, Mircea},
  eprint ={arXiv:2508.07494}, 
  year={2025},
  note = {Preprint}
}

@INPROCEEDINGS{farrage2025optimal,
  author={Farrage, Abdallah and Amir, Nur Azizah and Takahashi, Hideki and Sasai, Shintaro and Sakurai, Hitoshi and Okubo, Masaki and Uchiyama, Naoki},
  booktitle={2025 3rd International Conference on Mechatronics, Control and Robotics (ICMCR)}, 
  title={Optimal Trajectory Generation for Horizontal Boom Motion of Rotary Cranes Considering Load-Rotation Suppression}, 
  year={2025},
  volume={},
  number={},
  pages={113-117},
  doi={10.1109/ICMCR64890.2025.10963324}}

@article{shapira2007cranes,
  title={Cranes for building construction projects},
  author={Shapira, Aviad and Lucko, Gunnar and Schexnayder, Clifford J},
  journal={Journal of Construction Engineering and Management},
  volume={133},
  number={9},
  pages={690--700},
  year={2007},
  publisher={American Society of Civil Engineers},
  doi={https://doi.org/10.1061/(ASCE)0733-9364(2007)133:9(690)}
}

@article{aps2019mosek,
  title={Mosek optimization toolbox for matlab},
  author={ApS, Mosek},
  journal={User’s Guide and Reference Manual, Version},
  volume={4},
  number={1},
  pages={116},
  year={2019}
}

@article{burkhardt2023optimization,
  author={Burkhardt, Mark and Gienger, Andreas and Sawodny, Oliver},
  journal={IEEE Transactions on Control Systems Technology}, 
  title={Optimization-Based Multipoint Trajectory Planning Along Straight Lines for Tower Cranes}, 
  year={2024},
  volume={32},
  number={1},
  pages={290-297},
  doi={10.1109/TCST.2023.3308762}}

@article{bonnabel2020industrial,
  title={The industrial control of tower cranes: An operator-in-the-loop approach [applications in control]},
  author={Bonnabel, Silvere and Claeys, Xavier},
  journal={IEEE Control Systems Magazine},
  volume={40},
  number={5},
  pages={27--39},
  year={2020},
  publisher={IEEE},
  doi={10.1109/MCS.2020.3005256}
}

@article{markovsky2021behavioral,
  title={Behavioral systems theory in data-driven analysis, signal processing, and control},
  author={Markovsky, Ivan and D{\"o}rfler, Florian},
  journal={Annual Reviews in Control},
  volume={52},
  pages={42--64},
  year={2021},
  publisher={Elsevier},
  doi = {https://doi.org/10.1016/j.arcontrol.2021.09.005},
}

@article{kim2022data,
  title={Data-driven modeling and adaptive predictive anti-swing control of overhead cranes},
  author={Kim, Gyoung-Hahn and Yoon, Mahnjung and Jeon, Jae Young and Hong, Keum-Shik},
  journal={International Journal of Control, Automation and Systems},
  volume={20},
  number={8},
  pages={2712--2723},
  year={2022},
  publisher={Springer},
  doi={10.1007/s12555-022-0025-8}
}

@inproceedings{bao2020data,
  title={A Data-driven MPC algorithm for bridge cranes},
  author={Bao, HanQiu and An, Jing and Zhou, MengChu and Kang, Qi},
  booktitle={2020 International Conference on Advanced Mechatronic Systems (ICAMechS)},
  pages={328--332},
  year={2020},
  organization={IEEE},
  doi={10.1007/s11071-024-10384-6}
}

@article{chen2025data,
  title={Data-driven discrete learning sliding mode control for overhead cranes suffering from disturbances},
  author={Chen, Jianxun and Shi, Rong and Ouyang, Huimin},
  journal={Nonlinear Dynamics},
  volume={113},
  number={4},
  pages={3357--3372},
  year={2025},
  publisher={Springer}
  
}

@inproceedings{padoan2022behavioral,
  title={Behavioral uncertainty quantification for data-driven control},
  author={Padoan, Alberto and Coulson, Jeremy and Van Waarde, Henk J and Lygeros, John and D{\"o}rfler, Florian},
  booktitle={2022 IEEE 61st Conference on Decision and Control (CDC)},
  pages={4726--4731},
  year={2022},
  organization={IEEE},
  doi={10.1109/CDC51059.2022.9993002}
}

@book{markovsky2006exact,
  title={Exact and approximate modeling of linear systems: A behavioral approach},
  author={Markovsky, Ivan and Willems, Jan C and Van Huffel, Sabine and De Moor, Bart},
  year={2006},
  publisher={SIAM}
}


\end{document}